\documentclass[nohyperref]{article}

\usepackage[accepted]{icml2022/icml2022}

\usepackage{microtype}
\usepackage{graphicx}
\usepackage{subfigure}
\usepackage{booktabs}

\usepackage{hyperref}

\usepackage{xcolor}
\hypersetup{
    colorlinks,
    linkcolor={red!50!black},
    citecolor={red!50!black},
    urlcolor={red!50!black}
}

\usepackage{amsmath}
\usepackage{amssymb}
\usepackage{mathtools}
\usepackage{amsthm}

\usepackage{enumitem}
\usepackage{textcomp}

\usepackage[capitalize,noabbrev]{cleveref}

\usepackage[textsize=tiny]{todonotes}

\usepackage[title]{appendix}

\newcommand*\circled[1]{%
  \tikz[baseline=(C.base)]\node[draw,circle,inner sep=1.2pt,line width=0.2mm,](C) {#1};}
\newcommand*\Myitem{%
  \stepcounter{enumi}\item[\circled{\theenumi}]}
\makeatletter
\renewcommand*{\ALG@name}{Recipe}
\makeatother
\DeclareMathOperator*{\argmax}{arg\,max}

\theoremstyle{plain}
\newtheorem{theorem}{Theorem}[section]

\theoremstyle{definition}
\newtheorem{definition}[theorem]{Definition}

\theoremstyle{remark}

\renewcommand{\phi}{\varphi}

\newcommand{\R}{\mathbb{R}}

\newcommand{\sign}{\operatorname{sign}}

\newcommand{\norm}[1]{\Vert {#1} \Vert}

\begin{document}

\twocolumn[
\icmltitle{Investigating Generalization by Controlling Normalized Margin}

\icmlsetsymbol{equal}{*}

\begin{icmlauthorlist}
\icmlauthor{Alexander R. Farhang}{cal}
\icmlauthor{Jeremy Bernstein}{cal} 
\icmlauthor{Kushal Tirumala}{cal} 
\icmlauthor{Yang Liu}{comp}
\icmlauthor{Yisong Yue}{cal,comp}
\end{icmlauthorlist}
\icmlaffiliation{cal}{Caltech}
\icmlaffiliation{comp}{Argo AI}

\icmlcorrespondingauthor{Alexander R.\ Farhang}{afarhang@caltech.edu}
\icmlcorrespondingauthor{Jeremy Bernstein}{bernstein@caltech.edu}
\icmlcorrespondingauthor{Yisong Yue}{yyue@caltech.edu}

\icmlkeywords{generalization, optimization, neural networks, normalized margin, controlled studies}

\vskip 0.3in
]

\printAffiliationsAndNotice{}

\begin{abstract}
Weight norm $\|w\|$ and margin $\gamma$ participate in learning theory via the normalized margin $\gamma/\|w\|$. Since standard neural net optimizers do not control normalized margin, it is hard to test whether this quantity causally relates to generalization. This paper designs a series of experimental studies that explicitly control normalized margin and thereby tackle two central questions. First: \textit{does normalized margin always have a causal effect on generalization?} The paper finds that \textit{no}---networks can be produced where normalized margin has seemingly no relationship with generalization, counter to the theory of \citet{bartlett}. Second: \textit{does normalized margin ever have a causal effect on generalization?} The paper finds that \textit{yes}---in a standard training setup, test performance closely tracks normalized margin. The paper suggests a Gaussian process model as a promising explanation for this behavior.
\end{abstract}

\section{Introduction}
Despite significant progress, a complete explanation of the remarkable generalization capabilities of neural networks remains an open problem. Experimental studies often seek \textit{complexity measures} \citep{Prez2020GeneralizationBF} or optimization and architectural \textit{hyperparameters} \citep{Keskar2017OnLT} with explanatory power. But due to both the number of moving parts in a deep learning system and the cost of experimentation, unpacking the underlying effects is challenging.

One significant hurdle to a full scientific understanding of generalization is the presence of numerous potential \textit{confounders}. Even if a complexity measure, say, is strongly \textit{correlated} with generalization, this does not imply a causal link \citep{jiang2019fantastic}. Inspired by this observation, this paper singles out a quantity that is implicated by many theories as an important factor in generalization---normalized margin---and attempts to pin down its causal link to generalization. This quest is broken up into two separate sub-questions.

First, some theories suggest that normalized margin may have a very broad controlling effect on generalization \citep{bartlett}. To study that idea, this paper asks:
\begin{enumerate}[leftmargin=0.9cm,itemsep=0pt,topsep=0pt,label=\textbf{\textlangle Q\arabic*\textrangle}]
    \item \textit{\textbf{Does normalized margin always have a causal effect on generalization?}}\label{Q1}
\end{enumerate}
In other words: is a notion of normalized margin \textit{sufficient} to explain generalization? Is it the dominant factor? Or are there \textit{counterexamples}: settings where normalized margin is uninformative?

Second, some theories address the \textit{typical} behavior of function spaces rather than the \textit{worst case} \citep{mcallester1999some}. Perhaps normalized margin has a causal effect in these more typical settings. Consequently, this paper asks:

\begin{enumerate}[leftmargin=0.9cm,itemsep=0pt,topsep=0pt,label=\textbf{\textlangle Q\arabic*\textrangle}]
    \setcounter{enumi}{1}
    \item \textit{\textbf{Does normalized margin ever have a causal effect on generalization?}}\label{Q2}
\end{enumerate}

In other words: is a notion of normalized margin \textit{necessary} to build a complete picture of generalization? As posed, this question can be answered by finding \textit{positive examples} of settings where normalized margin has a causal effect.

To tackle these questions, this paper designs a series of experimental studies that take care to control both weight norms and margins of the learned predictors, resulting in the ability to target specific normalized margin distributions during training. The studies consider both \textit{spectrally-normalized} and \textit{Frobenius-normalized} margin distributions.

\textbf{In answer to \ref{Q1}, this paper finds that:}
\begin{enumerate}[leftmargin=0.875cm,itemsep=0pt,topsep=0pt]
    \item[\S~\ref{sec:snmb}] The effect that harder learning tasks correlate with smaller spectrally-normalized margin distributions---observed by \citet{bartlett}---can be reversed by controlling normalized margin distributions.
    \item[\S~\ref{sec:attack}] Pairs of networks can be found with \textit{very similar} Frobenius-normalized margin distributions but \textit{significantly different} generalization behavior.
\end{enumerate}

\textbf{In answer to \ref{Q2}, this paper finds that:}
\begin{enumerate}[leftmargin=0.875cm,itemsep=0pt,topsep=0pt]
    \item[\S~\ref{sec:em}] In a standard learning setting, controlling normalized margin does control generalization error.
\end{enumerate}
\textbf{Inspired by these findings, this paper further:}
\begin{enumerate}[leftmargin=0.875cm,itemsep=0pt,topsep=0pt]
    \item[\S~\ref{sec:gp-model}] Derives a  neural network--Gaussian process (NN--GP) model of the effect of normalized margin.
    \item[\S~\ref{sec:ensembles}] Finds that, in accordance with the NN--GP model, averaging the predictions of many small-normalized-margin networks improves their test error.
\end{enumerate}

\section{Related Work}
\label{sec:related}

\textbf{Support vector machines.} Normalized margin plays an important role in max-margin classifiers such as the support vector machine (SVM) \citep{cortes1995support, vapnik1999nature}. SVMs minimize weight norm at fixed margin, which is equivalent to maximizing margin at fixed weight norm. Learning theoretic arguments about the SVM have worked both via VC dimension \citep{boser_guyon_vapnik} and also via a PAC-Bayesian perspective \citep{why-svms-work}.

\textbf{Optimization procedures.}  Many machine learning techniques including soft-margin SVMs, Adaboost, and logistic regression employ margin maximizing loss functions \citep{rosset2003margin}. Notions of margin were often initially proposed as useful concepts for shallow models, and more recent work has extended these concepts to arbitrary layers of deep neural networks  \citep{elsayed2018large}. This includes both using the entire margin distribution, or just some of its statistics \citep{ jiang2018predicting}. Recent work has shown that, in certain problems, the gradient descent optimizer may be biased toward maximum normalized margin solutions without any explicit regularization \citep{soudry2018implicit}.

\textbf{Generalization bounds.} When bounding the risk of a learning algorithm, much of learning theory focuses on \textit{uniform convergence bounds} that hold for the worst function in a function class. This includes both VC bounds \citep{vapnik1999nature} and Rademacher bounds based on spectrally-normalized margin \citep{bartlett}. Another style of theoretical analysis known as PAC-Bayes theory \citep{langford2003pac, danroy_nonvacuous, neyshabur2017pac} focuses on the \textit{average} \citep{mcallester1999some} or \textit{typical} \citep{rivasplataKSS20,Prez2020GeneralizationBF} risk of functions in the function class.

Generalization bounds are often used to motivate \textit{complexity measures}---meaning formulae involving network properties that are intended to measure generalization ability \citep{neyshabur2015norm, jiang2019fantastic}. A fairly comprehensive survey of generalization bounds for neural networks is provided by \citet{Prez2020GeneralizationBF}.

\textbf{Experimental studies.} Researchers have found numerous puzzling empirical phenomena related to generalization in neural networks. Classic uniform convergence generalization bounds have been found to be vacuous in many realistic settings \citep{NEURIPS2019_05e97c20, zhang2021understanding}. Other effects such as \textit{double descent} of the population risk for increasing network width are of great interest \citep{Nakkiran2020Deep}. This has motivated a push towards greater empiricism in the study of generalization \citep{NEURIPS2018_a41b3bb3,extremeMemorization}. There have also been efforts to discover complexity measures that either correlate with \citep{nagarajan2019generalization, jiang2018predicting} or cause \citep{jiang2019fantastic, dziugaite2020search} generalization.

\section{Controlling Normalized Margin} 
\label{sec:quantities}

This section defines the \textit{normalized margin} of a neural network classification problem and develops a recipe to control this quantity. The recipe combines data normalization, a special loss function, and projected gradient descent.

\subsection{Defining Normalized Margin}

This subsection defines a notion of normalized margin in multi-layer perceptrons (MLPs), although the concept generalizes naturally to other network architectures.

The functional form of a depth-$L$ MLP is given by:
\begin{equation}\label{eq:mlp}
    f_L(x;w):= W_L \circ \phi \circ W_{L-1} \circ ... \circ \phi \circ W_1 (x),
\end{equation}
where $x\in\R^{d_0}$ is the \textit{input}, the matrices $w=(W_1,...,W_L)$ are the \textit{weights} and the elementwise function $\phi$ is the \textit{nonlinearity}. This paper will restrict to the ReLU nonlinearity $\phi(\cdot):=\max(0,\cdot)$, which is positive homogeneous. This means that the whole MLP is positive homogeneous of degree-$L$ in the weights $w$ and of degree-$1$ in the input $x$.

To use the network for binary classification, the output dimensionality of the network is set to 1 and the class decision is made via $x\mapsto\sign f_L(x;w)$. Then the margin of the network on input $x$ with binary label $y\in\{\pm1\}$ is given by:
\begin{equation*}
    \gamma(x, y;w) \coloneqq f_L(x;w) \cdot y.
\end{equation*}
To use the network for $k$-way classification, the output dimensionality of the network is set to $k$ and the class decision is made via $x\mapsto\argmax_i f_L(x;w)_i$. Then the margin of the network on input $x$ with label $y\in\{1,...,k\}$ is given by:
\begin{equation*}
    \gamma(x, y;w) \coloneqq  f_L(x;w)_y - \max_{i \neq y}  f_L(x;w)_i.
\end{equation*}
As defined, the margin inherits degree-$L$ homogeneity in the weights $w$ and degree-$1$ homogeneity in the input $x$ from the MLP. This is problematic, since weight and input rescalings affect neither classification decisions nor generalization performance. Therefore, it makes sense to define the \textit{normalized} margin $\overline{\gamma}_k$ for suitable ``norms'' $\norm{\cdot}_\star$ and $\norm{\cdot}_\dagger$:
\begin{equation*}
    \overline{\gamma}(x,y;w)\coloneqq\frac{\gamma_k(x,y;w)}{\norm{w}_\star \cdot \norm{x}_\dagger}.
\end{equation*}
The only real requirement on the ``norms'' is for $\norm{\cdot}_\star$ and $\norm{\cdot}_\dagger$ to be degree-$L$ and degree-$1$ positive homogeneous respectively. This paper will consider two such choices. The first is the most \textit{na\"{i}ve}:
\begin{definition}[Frobenius-normalized margin]\label{def:frob} Let $\norm{\cdot}_F$ denote the Frobenius norm, and let the $l$th weight matrix have dimension $d_l\times d_{l-1}$. The Frobenius normalized margin of training point $(x,y)$ is given by:
\begin{equation*}
    \overline{\gamma}_F(x,y;w)\coloneqq\gamma(x,y;w)\cdot\prod_{l=1}^L\frac{\sqrt{d_l}}{\norm{W_l}_F}\cdot \frac{\sqrt{d_0}}{\norm{x}_2}.
\end{equation*}
\end{definition}
The factors of dimension $d_l$ are included so that for standard weight and data scalings, the product term is of order one.

The second choice is a more involved notion of normalized margin that appears in a risk bound of \citet{bartlett}:
\begin{definition}[Spectrally-normalized margin]
\label{def:spectral_complexity}
Let $\norm{\cdot}_\sigma$ denote the spectral norm and $\norm{\cdot}_{2,1}$ denote the 1-norm of the column-wise 2-norm of a matrix. The spectrally-normalized margin of training point $(x,y)$ is given by:
\begin{equation*}
    \overline{\gamma}_\sigma(x,y;w)\coloneqq\gamma(x,y;w)\cdot\frac{1}{\mathcal{R}_w} \cdot \frac{\sqrt{d_0}}{\norm{x}_2},
\end{equation*}
where the \textit{spectral complexity} $\mathcal{R}_w$ is defined via:
    \begin{equation*}
        \mathcal{R}_w \coloneqq \left(\prod_{i=l}^{L} \norm{W_l}_{\sigma}\right) \left(\sum_{l=1}^{L} \frac{\norm{W_l^T - M_l^T}_{2, 1}^{2/3}}{\norm{W_l}_{\sigma}^{2/3}}\right)^{\!\!3/2}.
    \end{equation*}
    In this expression, $m=(M_1,...,M_L)$ are the weights of a reference network chosen before seeing the training data.
\end{definition}

The spectral complexity $\mathcal{R}_w$ matches Equation 1.2 of \citet{bartlett} after restricting to the ReLU nonlinearity, whose Lipschitz constant is one. The definition of spectrally normalized margin differs slightly in that \citet{bartlett} replace the factor of $\sqrt{d_0}/\norm{x}_2$ by $\norm{X}_F/n$ where $X$ is the training data matrix and $n$ is the number of training points. When each training point is normalized separately and $n$ is fixed---as in this paper's experiments---these definitions differ only by a constant factor.

\subsection{A Recipe for Controlling Normalized Margin}
In order to test the causal relationship between normalized margin and generalization, this section develops a recipe for directly controlling the distribution of Frobenius-normalized margins of a predictor over its training set (Recipe \ref{alg:control}). Due to the mathematical relationships between different norms, this also imposes a weak form of control over the spectrally-normalized margin distribution, which is exploited in \S~\ref{sec:snmb}.

The recipe has three steps: The first is to control the norm of each training input $\|x\|$. The second is to control the distribution of targeted margins $\gamma(x, y;w)$ across the training set. And the third is to control the product of Frobenius norms of the network weights $\prod_{l=1}^L \norm{W_l}_F$.

\begin{algorithm}[t]
\caption{Controlling Frobenius-normalized margin $\overline\gamma_F$. The recipe targets $\overline\gamma_F(x_i,y_i;w) = \alpha_i$ across training points $\{x_i,y_i\}_{i=1}^n$ for an $L$-layer MLP $f_L(x;w)$.}\label{alg:control}
\begin{enumerate}
    \Myitem Normalize all training inputs $x\in\R^{d_0}$ via:\vspace{-0.5em}
    \begin{flalign*}
        \qquad x &\gets x \cdot \frac{\sqrt{d_0}}{\norm{x}_2}.&& \vspace{-1em}
    \end{flalign*}
    \Myitem Set the loss function to:\vspace{-0.5em}
    \begin{flalign*}
        \qquad \mathcal{L}(w;\vec\alpha) &\gets \sum_{i=1}^n \left(f_L(x_i;w) - \alpha_i \cdot y_i\right)^2.&& \vspace{-1em}
    \end{flalign*}
    \Myitem After each descent step, normalize $W_l\in\R^{d_l\times d_{l-1}}$:
    \begin{flalign*}
        \qquad W_l &\gets W_l\cdot\frac{\sqrt{d_l}}{\norm{W_l}_F}, \text{ for layer } l=1,...,L.&&
    \end{flalign*}
\end{enumerate}
\end{algorithm}

\textbf{Step \circled{1}: controlling input norm.} The norm of each input can be controlled in a data pre-processing step. In all experiments, this paper controls the norm of each input $x\in\R^{d_0}$ by simply projecting the input on to the hypersphere of radius $\sqrt{d_0}$. For instance, $28\mathrm{px}\times28\mathrm{px}$ MNIST images are flattened into vectors and rescaled to have a 2-norm of $28$.

\textbf{Step \circled{2}: controlling margin distribution}. A special loss function $\mathcal{L}(w;\vec\alpha)$ is used to target networks with either a given margin or distribution of margins over the training set:
\begin{equation*}
        \mathcal{L}(w;\vec\alpha) \coloneqq \sum_{i=1}^n \left(f_L(x_i;w) - \alpha_i \cdot y_i\right)^2.
    \end{equation*}
For binary or one-hot labels $y_i$, minimizing this loss function to zero corresponds to returning a network with margin $\gamma(x_i, y_i;w)=\alpha_i$ on the $i$th training example. Setting all $\alpha_i$ to the same scalar $\alpha$ will be referred to as \textit{targeting margin $\alpha$}. This loss function is related to the ``rescaled square loss'' proposed by \citet{HuiSquareCrossEntropy}.
    
\textbf{Step \circled{3}: controlling product of weight norms.} Projected gradient descent is used to re-normalize the norm of each layer's weights after each iteration of network training. In practice, this paper employs the Nero optimizer \citep{pmlr-v139-liu21c}, which imposes a slightly stronger form of projection than is strictly necessary. In particular, Nero enforces that each row of every weight matrix has zero sum and unit length, so that $\norm{W_l}_F=\sqrt{d_l}$ as a consequence. The extra constraints are immaterial for the purposes of this paper, since the paper is only concerned with establishing examples and counterexamples of the causal effect of normalized margin---it may construct those examples in any fashion.

Combining all three steps yields Recipe \ref{alg:control}. Assuming that networks can be trained to zero loss, this recipe leads to exact control of the distribution of Frobenius-normalized margins (Definition \ref{def:frob}) over the training set. Because different norms weakly control each other---for instance:
\begin{align*}
    \norm{W_l}_F/\sqrt{\min(d_l,d_{l-1})} \leq \norm{W_l}_\sigma \leq \norm{W_l}_F,
\end{align*}
it follows that Recipe \ref{alg:control} also provides a weak form of control over the spectrally-normalized margin (Definition \ref{def:spectral_complexity}). This fact is exploited in \S~\ref{sec:snmb}.

\section{Normalized Margin is Insufficient to Explain Generalization}
\label{sec:sec_4}

The goal of this section is to tackle \ref{Q1}: \textit{Does normalized margin always have a causal effect on generalization?} 

The main finding of the section is that normalized margin can be decoupled from generalization performance. This includes a reversal of a previously reported correspondence between spectrally-normalized margin distributions and generalization in \S~\ref{sec:snmb}, and additional studies in \S~\ref{sec:attack} that decouple Frobenius-normalized margin from generalization performance. The experiments in \S~\ref{sec:attack} are referred to as \textit{twin network studies} since they produce pairs of networks with very similar Frobenius-normalized margin distributions but significantly different test performance. These results constitute counterexamples suggesting that normalized margin alone cannot causally explain generalization.

\begin{figure}
\centering
\hspace{3.2em}
\begin{tikzpicture}[->,>=stealth,auto,node distance=1.5cm,thick]
    \node (1) {};
    \node (2) [right of=1] {};
    \draw [<->] (2.south) to [out=150,in=30] (1.south) node[midway,above,xshift=0.75cm,yshift=0.1cm] {order reversed};
\end{tikzpicture}\vspace{-5pt}
\includegraphics[width=\columnwidth]{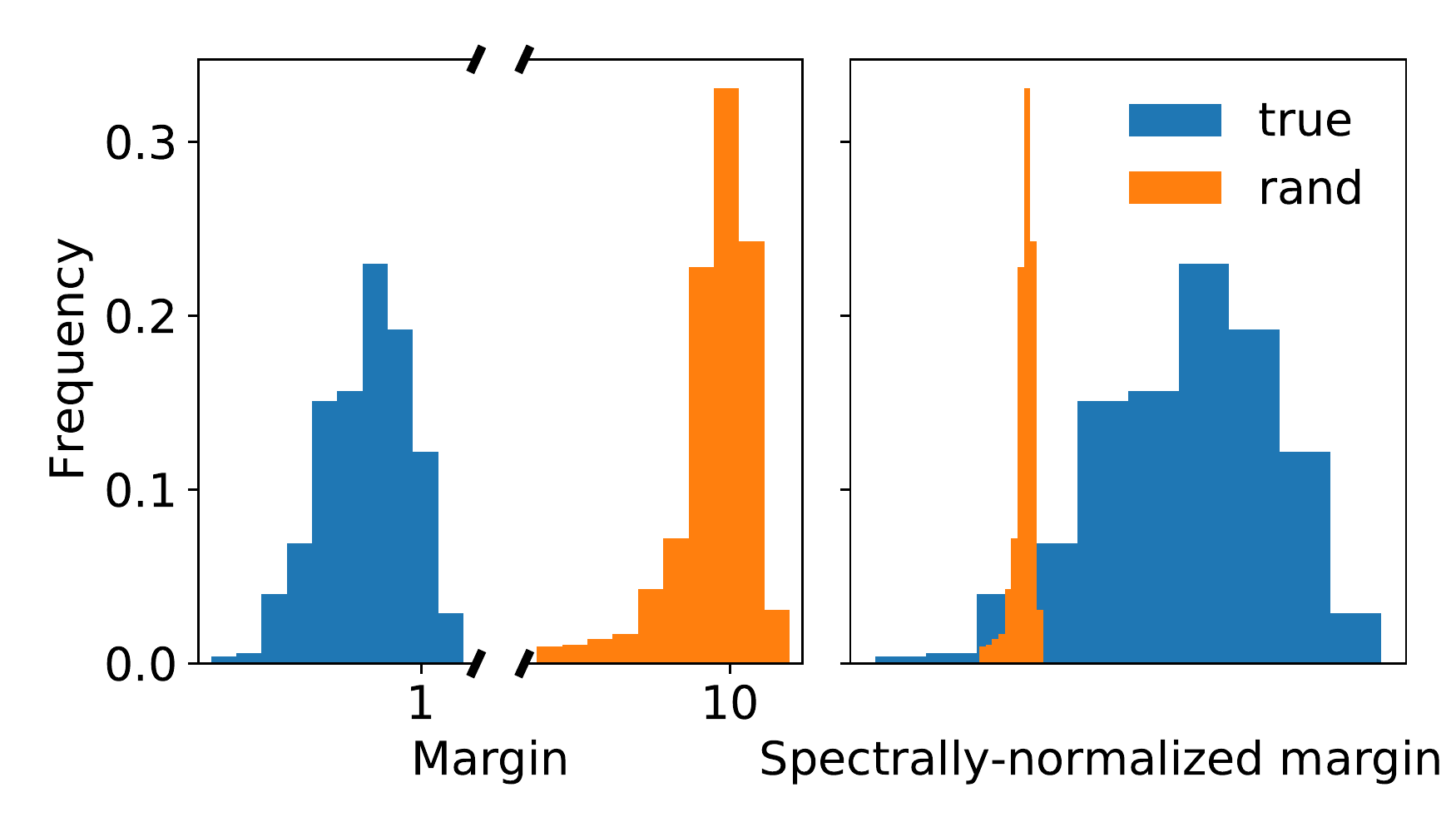}\vspace{-10pt}
\caption{Reproducing the effect of \citet{bartlett}. \textit{True labels} corresponds to a standard learning problem with a generalizing solution (81\% test accuracy), while \textit{random labels} corresponds to an impossible learning problem with a non-generalizing solution (8.3\% test accuracy). After spectral-normalization, the margin distribution of the generalizing solution falls above that of the non-generalizing solution.}
\label{fig:bartlett_reproduce}
\end{figure}
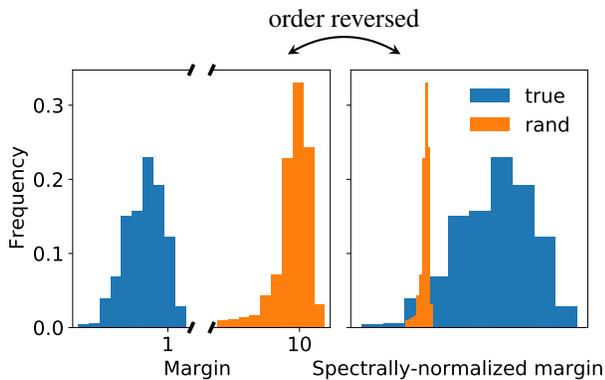

\subsection{Reversing Spectrally-Normalized Margin Bounds}
\label{sec:snmb}

This section shows that, through control, spectrally-normalized margin can be made to both correlate \textit{and} anti-correlate with generalization error.  These results are motivated by a prior finding that a measure of spectrally-normalized margin derived from a theoretical bound can correlate with generalization ability \citep{bartlett}. This section's results highlight the risk in inferring a causal connection from a correlational study.

\textbf{Background.} Spectrally-normalized margin distributions have been proposed as a promising method to understand generalization in neural networks \citep{bartlett}. The theory is derived from a risk bound related to Definition \ref{def:spectral_complexity}. In spirit, this bound is given by:
\begin{align}
    R(f) \lessapprox 
    \widehat{R}_\gamma(f)+ \frac{\|X\|_F \mathcal{R}_A}{\gamma n}, \label{eq:bartlett_bound}
\end{align}
where $R(f)$ is the population risk, $\widehat{R}_\gamma(f)$ measures what \citet{bartlett} refer to as the sample ``ramp loss'' at margin $\gamma$, $\|X\|_F$ is the Frobenius norm of the training data matrix and $n$ is the number of training samples.

\citet{bartlett} also provide a graphical way to understand the bound via the relative placement of margin distributions. In particular, for ``any fixed point on the horizontal axis, if the cumulative distribution of one density is lower than the other, then it corresponds to a lower right hand side" of their bound. For the purposes of this paper, this means that if one learner's spectrally normalized margin distribution lies fully above that of a second learner, then the first learner should generalize better according to \citet{bartlett}'s theory.

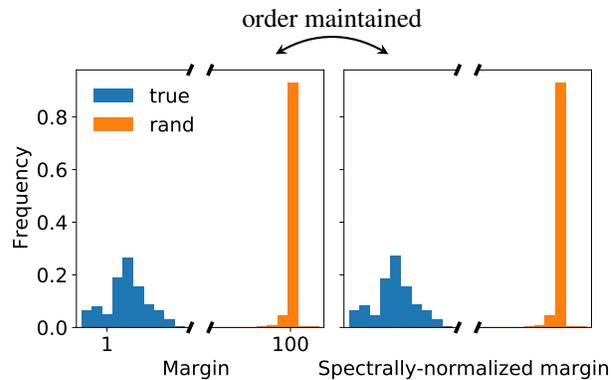
\begin{figure}
\centering
\hspace{2em}
\begin{tikzpicture}[->,>=stealth,auto,node distance=1.5cm,thick]
    \node (1) {};
    \node (2) [right of=1] {};
    \draw [<->] (2.south) to [out=150,in=30] (1.south) node[midway,above,xshift=0.75cm,yshift=0.1cm] {order maintained};
\end{tikzpicture}\vspace{-5pt}
\includegraphics[width=\columnwidth]{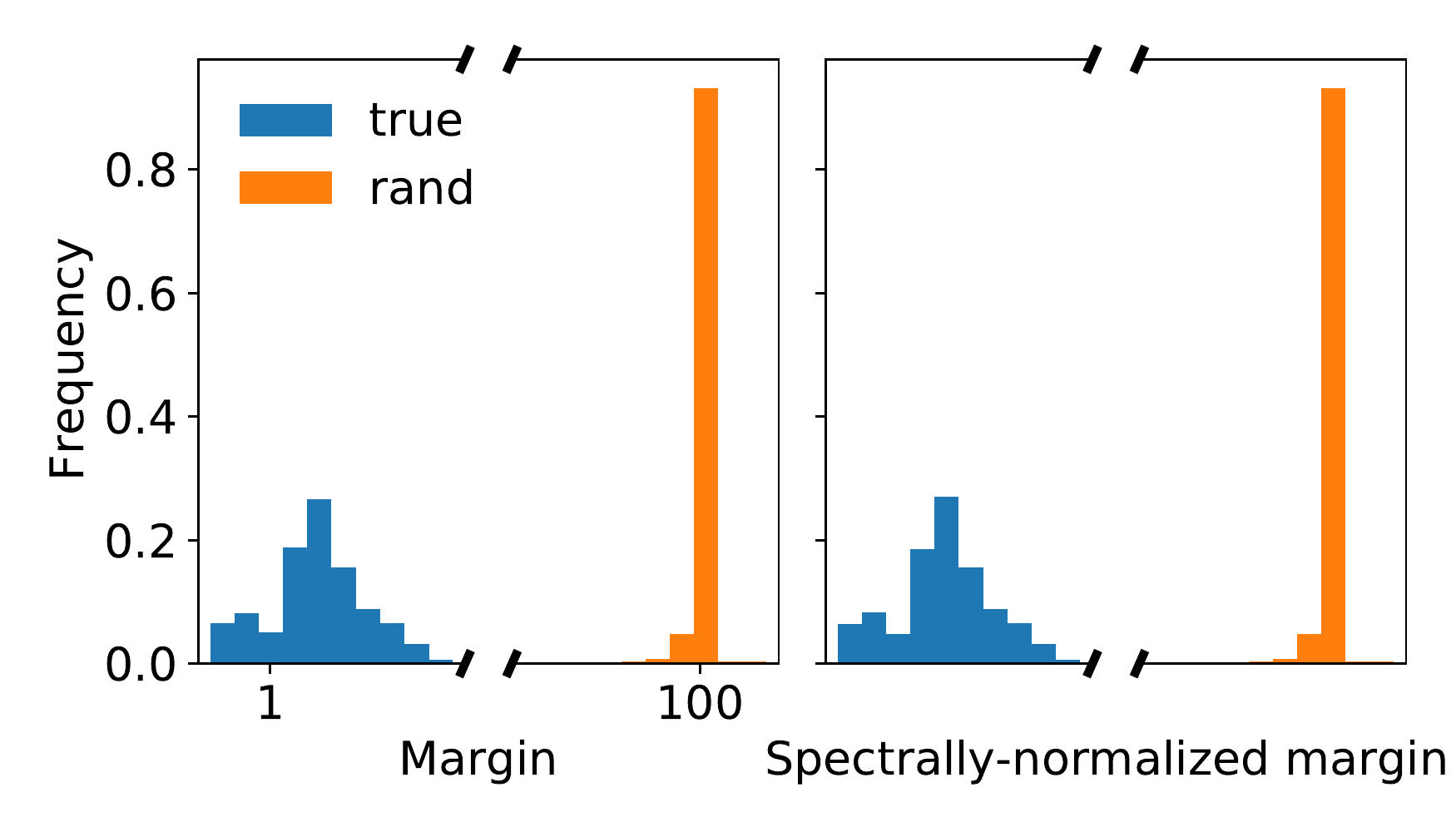}\vspace{-10pt}
\caption{Breaking the effect of \citet{bartlett}. This figure is under the same experimental setting as Figure \ref{fig:bartlett_reproduce}, except Recipe \ref{alg:control} has been used to greatly inflate the margin distribution on the \textit{random label} task through controlled optimization. The ordering of the spectrally-normalized margin distributions no longer reflects the generalizability of the corresponding solutions (\textit{true labels}: 81\% test accuracy vs \textit{random labels}: 10\% test accuracy)}
\label{fig:bartlett_placeholder}
\end{figure}

Such a graphical comparison is conducted in Figure \ref{fig:bartlett_reproduce}. In this case, spectral normalization successfully reorders two margin distributions of correctly classified points with correct or random labels such that the generalizing network's distribution places most of its mass to the right of the non-generalizing network.  %
Given this result, it is tempting to surmise that spectrally-normalized margin may be a dominant causal factor in a network's generalization ability. The experiments in this section explore this hypothesis.

\begin{figure}
\centering
\includegraphics[width=\columnwidth]{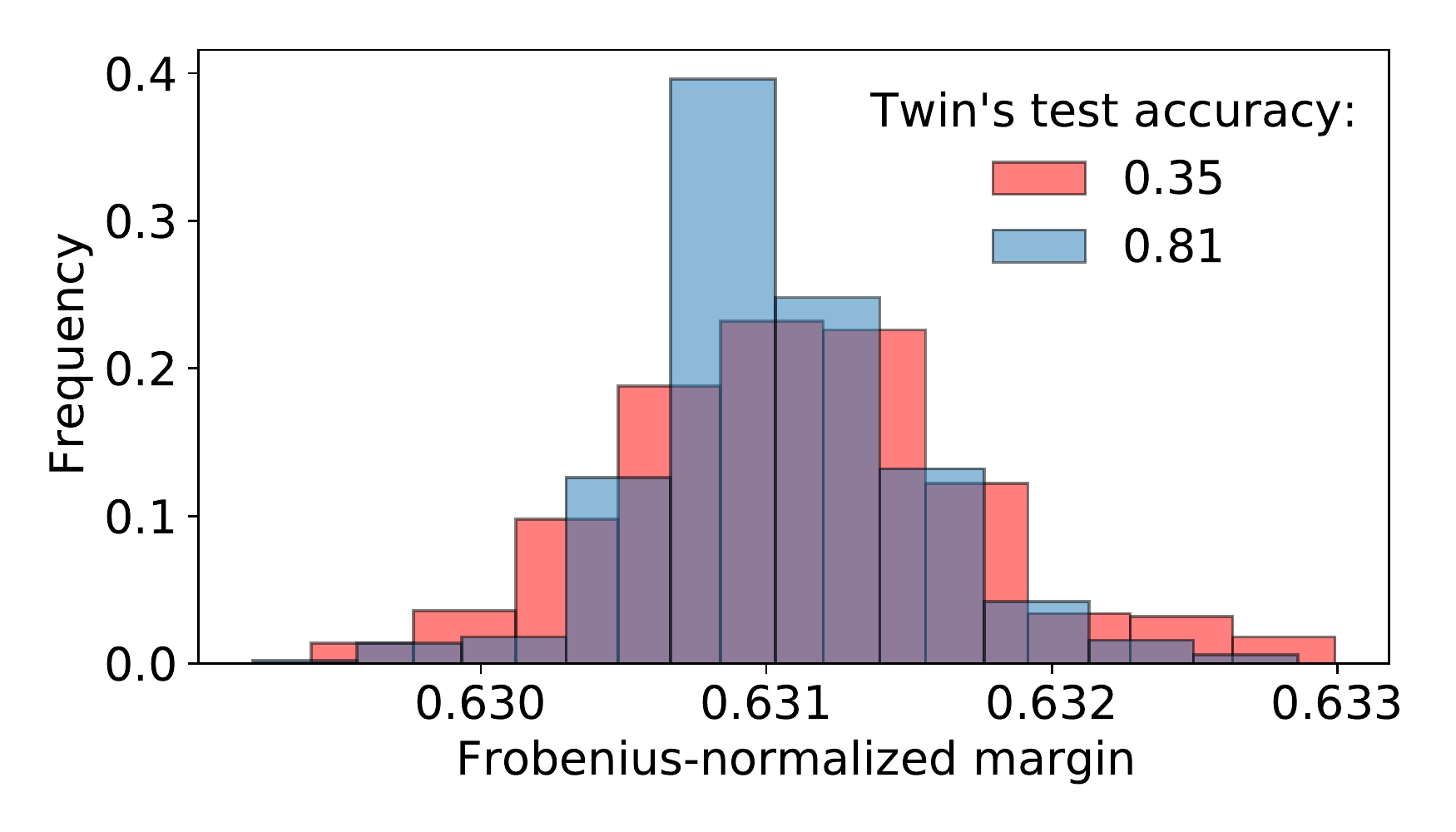}
\vspace{-25pt}
\caption{``Twin networks'' with overlapping Frobenius-normalized margin distributions but significantly different test performance. The poorly generalizing network was selected via the attack set method of \citet{Wu2017TowardsUG}, while its twin used standard training.}
\label{fig:augment_hist_placeholder}
\end{figure}

\textbf{Experiments.}
Two sets of experiments were performed, each of which trained two MLPs  on $1000$ point subsets of MNIST to classify either true or randomly labeled data for $10$-class classification.  Using full-batch gradient-based optimization, a scaled squared loss function, and data normalization described in \S~\ref{sec:quantities}, two networks with identical architecture were trained on either true or random labels.  Note that only networks trained on true labels can possibly generalize.  The experiments in this section varied the targeted label scale $\alpha$, forcing the networks' margin distributions to converge to $\alpha$.  Networks trained on true labels always target $\alpha = 1$.  
\begin{itemize}[topsep=0pt,itemsep=0pt]
    \item \textbf{Experiment 1:} Spectrally-normalized margin distributions correspond with generalization ability in networks trained without weight constraints, with random labels targeting $\alpha = 10$ (Figure \ref{fig:bartlett_reproduce}).
    \item \textbf{Experiment 2:} Spectrally-normalized margin distributions do not correspond with generalization ability in networks trained with Frobenius weight constraints, with random labels targeting $\alpha = 100$ (Figure \ref{fig:bartlett_placeholder}).
\end{itemize}

\textbf{Findings.}  
The two experiments described above show that spectrally-normalized margin distributions do not track a network's generalization power.  
By running controlled experiments, Figures \ref{fig:bartlett_reproduce} and \ref{fig:bartlett_placeholder} show opposing correspondences between spectrally-normalized margin and generalization ability.
In other words, whereas Figure \ref{fig:bartlett_reproduce} is consistent with the generalization bound and uncontrolled empirical study in \citet{bartlett}, Figure \ref{fig:bartlett_placeholder} shows the opposite effect.  %
Overall, this study suggests that spectrally-normalized margin alone does not causally control generalization.

\begin{figure}
\centering
\includegraphics[width=\columnwidth]{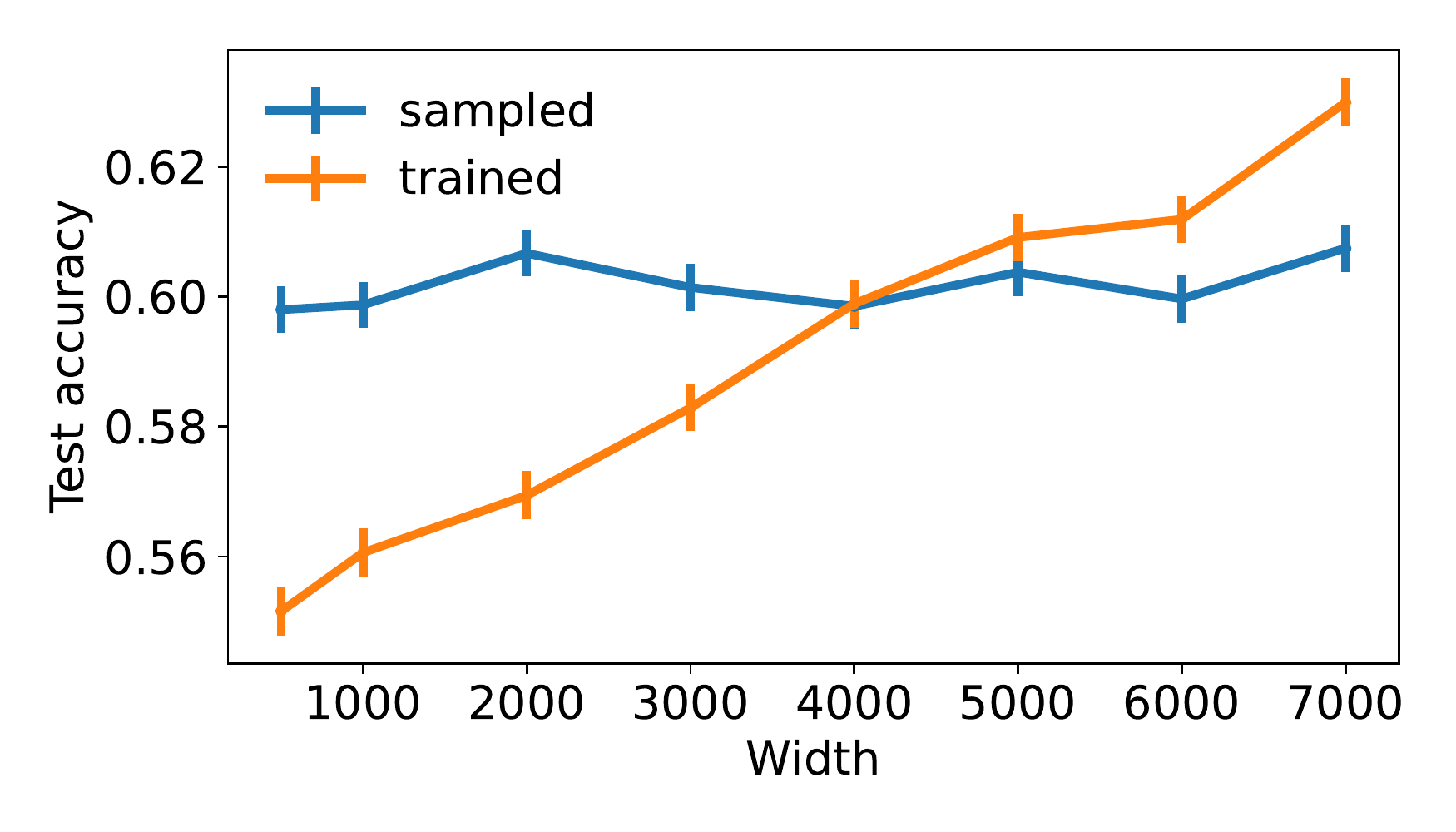}
\vspace{-25pt}
\caption{``Twin networks'' with nearly identical Frobenius-normalized margin distributions that were selected via different optimization procedures. Each \textit{sampled} network was found via rejection sampling, while its \textit{trained} twin used gradient-based optimization to match its margin distribution. To make rejection sampling feasible, a very small learning problem involving 5 MNIST samples was used. At each network width, mean test accuracy and standard error of the mean is reported for $1000$ pairs of twins. For near identical normalized margin distribution, the different training procedures led to different test accuracies.}
\label{fig:sampled_v_trained}
\end{figure}

\subsection{Twin Network Studies}
\label{sec:attack}

To further explore the sufficiency of normalized margin in explaining generalization ability, this section designs two \textit{twin network} studies to control Frobenius-normalized margins and observe their effect on generalization performance.

\textbf{Background.} In order to find neural network solutions with varying generalization performance, recent studies have included training points that are incorrectly labeled to reduce test performance \citep{Wu2017TowardsUG, zhang2021flatness}.  Inspired in part by these approaches, these experiments sought to produce networks that can produce similar normalized margin distributions with varying test performance through the inclusion of an \textit{attack set} of training points with random labels (\textbf{Experiment 1}).  

Previous work has studied the implicit biases of both gradient-descent \citep{soudry2018implicit} and randomly sampling parameters \citep{valle_perez2018deep, de2019random}. In the infinite-width limit, the posterior distribution over the function space is similar between networks trained by SGD or random sampling \citep{mingard2021sgd}.  \textbf{Experiment 2} in this section fixes normalized margin between twin networks that differ in training method (random sampling vs. gradient-based optimization) and analyzes the difference in generalization performance.

\textbf{Experiment 1: Attack set twin.} Twin networks were trained on identical subsets of $500$ points of the MNIST training set that targeted the same normalized margins.
However, one of the networks was also trained to target the same normalized margin for additional $1000$ randomly labeled train points, the attack set.  Both networks are MLPs with identical architectures under Frobenius weight norm $\|w\|$, targeted margin $\gamma$, and data norm $\|X\|$ control as specified in Recipe \ref{alg:control}.

\textbf{Experiment 2: Rejection sampled twin.} Pairs of networks with identical architectures (see Appendix \ref{sec:twin_network_exp_details}) are generated to have matching Frobenius-normalized margins on a training set, but are obtained via different optimization methods (sampling vs. gradient-based optimization).  First, a network $f_{\text{sampled}}$ is found by randomly sampling Frobenius-norm constrained networks until a small training set of binary MNIST data is perfectly classified.  Then, a second twin $f_{\text{trained}}$ is trained using Frobenius constrained gradient descent with Nero to perfectly mimic the output of $f_{\text{sampled}}$ on the same training set, resulting in two networks with matched normalized margins. This procedure is repeated $1000$ times.

\textbf{Findings.}  Networks can have  similar Frobenius-normalized margin distributions while exhibiting drastically different generalization.  
The results from the attack set experiments, \textbf{Experiment 1}, in Figure \ref{fig:augment_hist_placeholder} show the Frobenius-normalized margin distributions on the correctly labeled data points each of the twins was trained on.  Though the twin networks trained with (red) or without (blue) the addition of an attack set have similar normalized margin distributions, they have substantially different test performance ($35\%$ vs. $81\%$ accuracy).  The normalized margin distribution's placement can be somewhat arbitrarily controlled irrespective of generalization ability for attack set twins by targeting various margin scales $\alpha$ in the scaled loss function.  These results suggest that neural networks could have matching normalized margin distributions and thus similar functional output on the train set, yet one could display pathologically reduced generalization. %

There is also a gap in generalization performance between twin networks from \textbf{Experiment 2}, which have nearly identical normalized margins but were trained with different optimization methods.
As shown in Figure \ref{fig:sampled_v_trained}, for MNIST 0 vs. 1 classification, some architectures (i.e some fixed widths) exhibit significantly different generalization performance between $f_{\text{sampled}}$ and $f_{\text{trained}}$. This effect is observed across random seeds and different learning tasks for binary classification in MNIST and CIFAR-10 (see Appendix \ref{sec:twin_network_exp_details}).
This difference in generalization performance between $f_{\text{sampled}}$ and $f_{\text{trained}}$ cannot be attributed to normalized margin, since both margin $\gamma$ and weight norms $\|w\|$ are nearly identical across the two networks, and instead may be due to the implicit biases of the corresponding optimization methods.

\section{Normalized Margin May Be  Necessary to Explain Generalization}
\label{sec:sec_5}
The goal of this section is to tackle \ref{Q2}: \textit{Does normalized margin ever have a causal effect on generalization?} 

While \S~\ref{sec:sec_4} presented multiple settings where normalized margin does not causally impact generalization, this section seeks the opposite: settings where normalized margin does causally effect generalization.

\subsection{Normalized Margin in Standard Training}
\label{sec:em}
\begin{figure}
\centering
\includegraphics[width=\columnwidth]{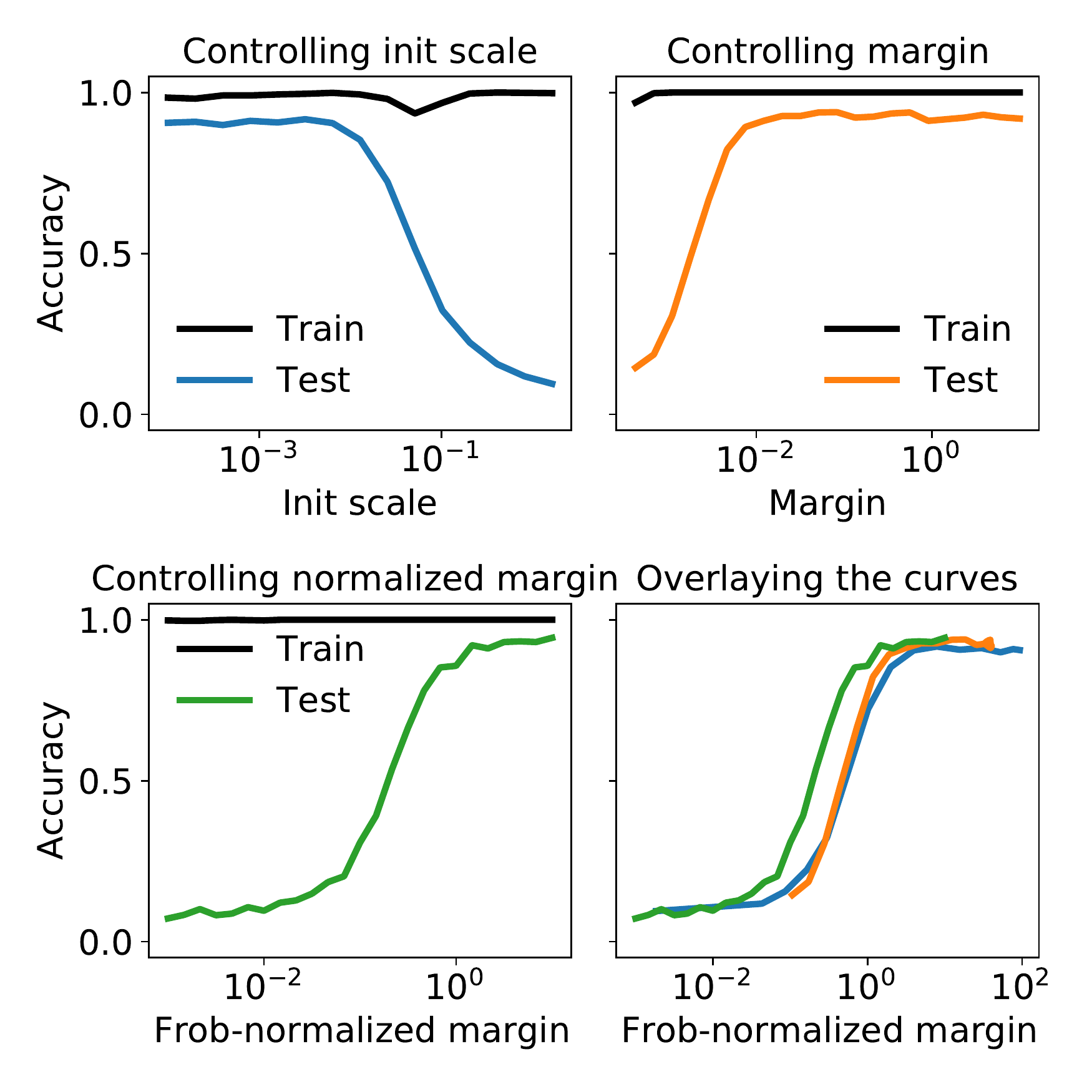}
\vspace{-25pt}
\caption{Accuracy as a function of controlling initialization scale, margin, or Frobenius-normalized margin.  The last plot overlays the test accuracy from the other three plots as a function of each network's targeted margin divided by the layer-wise product of Frobenius norms. The close overlap suggests that Frobenius-normalized margin may largely explain the behavior in the first two plots.}
\label{fig:em_placeholder}
\end{figure}

This section explores the effect of normalized margin in networks trained in a more benign manner than was considered in \S~\ref{sec:sec_4}.  Three experiments are conducted, each using a different subset of control presented in \S~\ref{sec:quantities}.  They all provide evidence supporting the idea that larger targeted normalized margins correspond with a network's increased test performance (bottom right panel of Figure \ref{fig:em_placeholder}).

\textbf{Background.}  A recent study observed how a neural network's \textit{scale of initialization} can tightly control its generalization ability \citep{extremeMemorization}.  In particular, by varying the scale of initialization of the first layer, one could cause a network to interpolate between good and chance test performance in the extreme case.  This phenomenon may have connections with how overparameterized networks can operate in regimes known as ``kernel'' or ``rich'' depending on the model's similarity to kernel regression throughout learning \citep{woodworth20a, Geiger_2020}.  But notably a network's scale of initialization can affect the scale of its weight norms and thus its normalized margin.

\textbf{Experiments.}
Controlled experiments were designed to understand this phenomenon from the perspective of normalized margin.  2-layer MLPs were trained for $10$-class classification on $1000$ point subsets of MNIST. %
The following three experiments were run:
\begin{enumerate}[topsep=0pt,itemsep=0pt]
    \item The initialization scale was varied, while the target margin was fixed to 1.
    \item The targeted margin was varied, for a fixed initialization scale and with weight projection removed.
    \item Frobenius-normalized margin was directly controlled and varied using Recipe \ref{alg:control}.
\end{enumerate}

\textbf{Findings.} 
These experiments reveal a correspondence between a network's generalization performance and its Frobenius-normalized margin; for a given network that can generalize, it tends to generalize better when it targets a larger normalized margin.  Figure \ref{fig:em_placeholder} demonstrates that generalization can be controlled by targeting certain normalized margins. 
The bottom right panel of Figure \ref{fig:em_placeholder} registers all of the test performance curves onto the same scale, as a function of their normalized targeted margin.  By constraining the layerwise Frobenius norms  and targeting specific margins with a modified squared loss function, generalization ability can be controlled by varying a network's targeted margin.  Whereas the scale of initialization is set at the beginning of training and then free to vary during the dynamics of optimization, the targeted margin remains constant throughout training.  These experiments suggest that in a standard training setting for networks that generalize, controlling normalized margin does control generalization.

\section{Building on the Controlled Studies}
So far, this paper has made two main findings. First, normalized margin seems insufficient to fully explain generalization. \S~\ref{sec:sec_4} showed that \textit{through careful control} one can break a reported link between normalized margin and generalization. Second, normalized margin does seem to have a strong controlling effect on generalization in less adversarial situations, as shown in \S~\ref{sec:sec_5}. The goal of this section is to develop a model that is consistent with these findings and has predictive power over the effect of normalized margin.

In particular, \S~\ref{sec:gp-model} constructs a model of normalized margin based on the NN–GP correspondence. \S~\ref{sec:ensembles} points out that this model makes concrete predictions about the behavior of \textit{ensembles} of small-normalized-margin networks. These predictions are tested and verified---providing promising evidence in favor of the Gaussian process model.

\subsection{A Gaussian Process Model of Normalized Margin}
\label{sec:gp-model}

A Gaussian process (GP) can in principle, up to certain technical conditions, fit any function. Therefore a GP should be able to represent functions of arbitrary margin that behave arbitrarily badly away from the training data. Since GP classification is effective in practice \citep{gpml}, such poorly behaving functions must not be selected for by GP inference. To test whether this is essentially the same behavior that is being observed in \S~\ref{sec:sec_4} and \S~\ref{sec:sec_5}, one needs to build a model of GP classification that explicitly involves a normalized margin parameter. 

This section accomplishes that task via the \textit{neural network--Gaussian process correspondence} (NN-GP) \citep{radford, lee2018deep, g.2018gaussian}. Consider an $L$-layer ReLU-MLP (Equation \ref{eq:mlp}) with weight matrices $w=(W_1,...,W_L)$, where the $l$th layer's weight matrix $W_l$ has dimension $d_l\times d_{l-1}$. Consider randomly sampling the weights at each layer according to $W_l^{(ij)}\overset{\mathrm{iid}}{\sim}\mathcal{N}(0,\sigma^2/d_{l-1})$, where the parameter $\sigma$ sets the prior scale of each layer.

Sending the layer widths $d_1,...,d_L$ to infinity, the NN--GP correspondence states that the distribution of network outputs under this prior on the weights is given by a GP with mean zero and covariance function \citep{lee2018deep}:
\begin{align*}
    \Sigma(x,x^\prime) = \sigma^{2L}\cdot\underbrace{h \circ ... \circ h}_{L-1 \text{ times}}\left(\frac{x^T x^\prime}{d_0}\right),
\end{align*}
where $h(t)\coloneqq\tfrac{1}{\pi}\cdot [ \sqrt{1-t^2} + t\cdot (\pi - \arccos t)].$ This is the \textit{compositional arccosine kernel} of \citet{choandsaul}.

To construct the posterior distribution over a test point $x$ given a training set $X=\{x_1,...,x_n\}$ and a vector of binary training labels $Y\in\{\pm1\}^n$, one requires the \textit{Gram matrix} $\Sigma_{XX}$, \textit{Gram vector} $\Sigma_{xX}$ and \textit{Gram scalar} $\Sigma_{xx}$ defined by:
\begin{align*}
    \Sigma_{XX}^{(ij)} \coloneqq \Sigma(x_i,x_j); \;
    \Sigma_{xX}^{(i)} \coloneqq \Sigma(x,x_i); \;
    \Sigma_{xx} \coloneqq \Sigma(x,x).
\end{align*}
This paper also defines the \textit{normalized Gram tensors} via: \begin{align*}
    \widehat\Sigma_{XX} \coloneqq \frac{\Sigma_{XX}}{\sigma^{2L}}; \;\;\;\;\;\;
    \widehat\Sigma_{xX} \coloneqq \frac{\Sigma_{xX}}{\sigma^{2L}}; \;\;\;\;\;\;
    \widehat\Sigma_{xx} \coloneqq \frac{\Sigma_{xx}}{\sigma^{2L}}.
\end{align*}

Consider scaling up the training labels by a \textit{margin parameter} $\gamma$. The distribution over functions that interpolate the training points $(X,\gamma Y)$ evaluated at test point $x$ is then:
\begin{align*}
    &\mathcal{N}\left(\gamma\cdot\Sigma_{xX}\Sigma_{XX}^{-1}Y,  \Sigma_{xx}-\Sigma_{xX}\Sigma_{XX}^{-1}\Sigma_{Xx}\right) \\
    &= \mathcal{N}\big(\gamma\cdot\underbrace{\widehat\Sigma_{xX}\widehat\Sigma_{XX}^{-1}Y}_{\eqqcolon C_1}, \sigma^{2L} \cdot\underbrace{\widehat\Sigma_{xx}-\widehat\Sigma_{xX}\widehat\Sigma_{XX}^{-1}\widehat\Sigma_{Xx}}_{\eqqcolon C_2}\big).
\end{align*}
The signal-to-noise ratio of this GP posterior is set by the parameter $\gamma / \sigma^L$. Since $\gamma$ sets the scale of the outputs, and $\sigma$ sets the prior scale of each layer's weights, $\gamma / \sigma^L$ has an interpretation as the \textit{normalized margin} of the posterior. The next section tests the effect of $\gamma / \sigma^L$ on generalization.

\begin{figure}
\centering
\includegraphics[width=\linewidth]{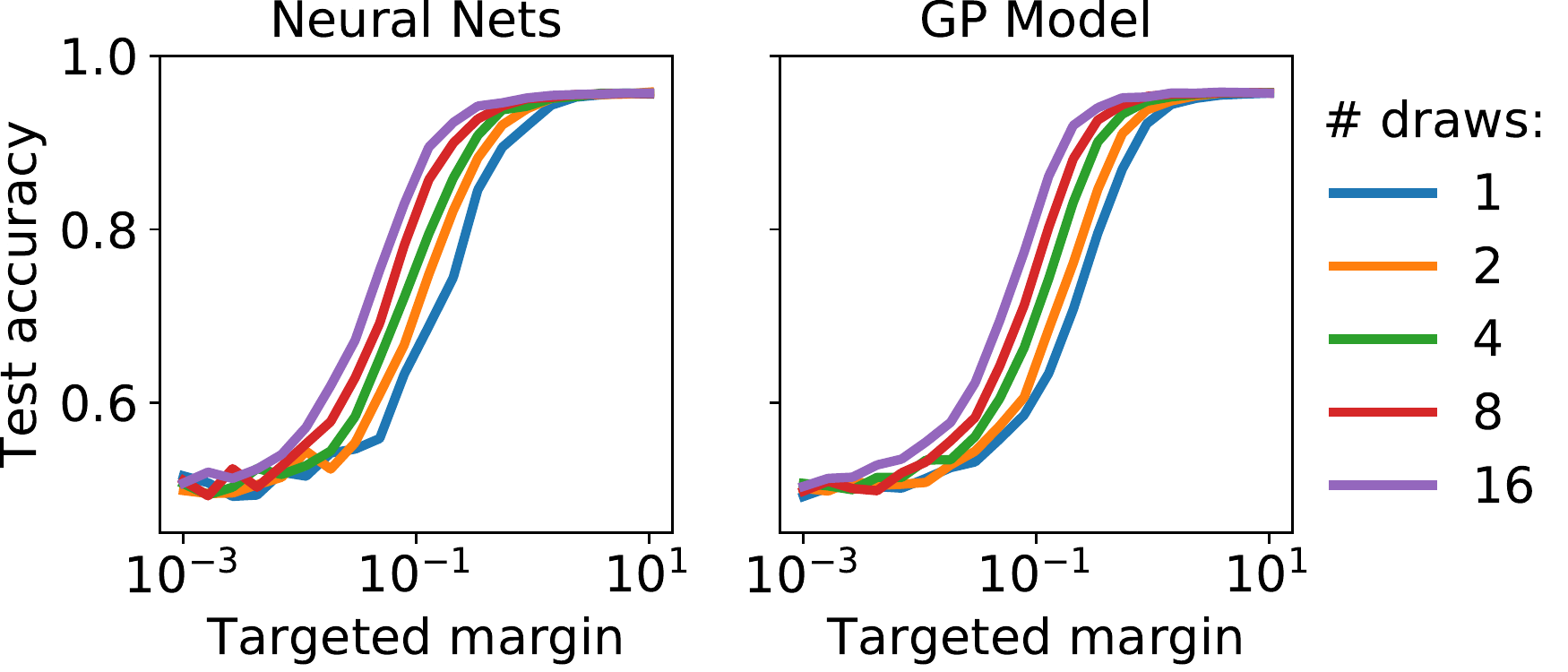}
\vspace{-20pt}
\caption{Averaging the predictions of many NNs (left) and NN--GP posterior samples (right) as a function of normalized margin. For NNs, \textit{\# draws} refers to the number of networks trained by Nero. \textit{Targeted margin} refers to \textit{Frobenius-normalized margin} for NNs and $\gamma/\sigma^L$ (see Equation \ref{eq:gp-ensemble}) for NN--GP draws. Test performance increases in a very similar way for both NNs and NN--GP draws, as a function of both ensemble size and normalized margin.}
\label{fig:ensembles}
\end{figure}
\subsection{Ensemble Behavior of Normalized Margin}
\label{sec:ensembles}

This section studies the NN--GP model of normalized margin developed in \S~\ref{sec:gp-model}. The central prediction of the model is shown to be that \textit{averaging small-normalized-margin functions} should have the same effect on generalization as \textit{increasing the normalized margin}. This effect is found to map back to finite width MLPs in Figure \ref{fig:ensembles}.

\textbf{Background.} In \S~\ref{sec:gp-model}, it was shown that the NN--GP predictive distribution at target margin $\gamma$ and layer scale $\sigma$ is:
\begin{equation*}
    \mathcal{N}(\gamma \cdot C_1, \sigma^{2L} \cdot C_2),
\end{equation*}
where $C_1$ and $C_2$ are independent of $\gamma$ and $\sigma$. Also, the average of $m$ iid draws from this distribution follows:
\begin{equation}
    \mathcal{N}(\gamma \cdot C_1, \sigma^{2L} \cdot C_2 / m). \label{eq:gp-ensemble}
\end{equation}
Since the variance of the posterior corresponds to adding Gaussian noise to the predictions, it is only reasonable that this variance should harm prediction quality. To force the posterior to concentrate on its mean, one may either:
\begin{enumerate}[label=\alph*), topsep=0pt, itemsep=0pt]
    \item Let the normalized margin $\gamma/\sigma^L\to\infty$;
    \item Let the number of ensemble members $m\to\infty$.
\end{enumerate}
This model may be tested for finite width NNs simply by replacing $\gamma/\sigma^L$ with the Frobenius-normalized margin.

\textbf{Experiments.}  Finite width MLPs were trained on subsets of MNIST for even/odd classification using layerwise Frobenius control and margin control as prescribed in Recipe \ref{alg:control}.  Each individual model in a given ensemble is trained on the same subset of data and their output activations are averaged and then binarized to form a prediction.  This is performed over a range of targeted Frobenius-normalized margins.  The same experiment was repeated for NN--GP draws using the ensembled predictive distribution given in Equation \ref{eq:gp-ensemble}.

\textbf{Findings.}  For both ensembles of networks and GP draws, test performances increases as a function of both ensemble size and targeted normalized margin. As shown in Figure \ref{fig:ensembles}, individual large-margin classifiers attain the same test accuracy as an ensemble-average of small-margin classifiers. The functional form of the curves for finite width NNs and NN--GP draws is remarkably similar.
\section{Discussion}
The paper has presented a set of controlled experiments to address the sufficiency of normalized margin to explain generalization in \textit{all settings}, and its necessity in \textit{typical settings}.  The counterexamples in \S~\ref{sec:sec_4} show that spectrally- and Frobenius-normalized margin are not sufficient to explain generalization performance in general, since for instance normalized margin distributions can be somewhat arbitrarily inflated through controlled optimization without yielding good generalization performance. However, the positive examples in \S~\ref{sec:sec_5} demonstrate that normalized margin can control test performance in less adversarial settings.

This section discusses three topics: first, how the paper relates to the pursuit of a more scientific understanding of deep learning; second, the potential for the paper's results and techniques to inform future developments in learning theory; and third, a possible application of normalized margin control to uncertainty quantification via deep ensembles.

\subsection{Predictive Models and Controlled Studies}
At the core of the scientific process is the construction of predictive theories and models, which are in turn used to generate and test falsifiable hypotheses via experiment. Controlled studies are often considered the ``gold standard'' in experimental design for this kind of hypothesis testing. Under this light, this paper has developed a means of controlling normalized margin in order to test a generalization theory based on spectrally normalized margin distributions (\S~\ref{sec:snmb}). Finding this theory wanting, the paper constructed a new model based on a notion of normalized margin in Gaussian processes (\S~\ref{sec:gp-model}). This new model was found to yield accurate predictions about the behavior of ensembles of neural networks.

A model is a simplification or abstraction of a system that throws away the messy details while attempting to capture the system's essence.  Models are important for their ability to reveal insights and relationships about a system that are difficult to see directly.  The NN--GP model of normalized margin proposed in \S~\ref{sec:gp-model} suggests that there are still valuable insights that can be drawn from established models. In this instance, the NN--GP model can provide insight about generalization in neural networks by focusing on the function space prior without appealing to more involved models such as the \textit{neural tangent kernel} \citep{NTKjacot}.

In contrast to the type of controlled experimentation conducted in this paper, much work studies phenomena in deep learning without experimental intervention on the objects of theoretical interest \citep{bartlett}. Other work moves toward a greater level of control---for instance \citet{jiang2019fantastic} control optimization hyperparameters and subsequently attempt to tease out causal relationships between complexity measures and generalization.  However, this paper goes a step further by directly intervening on the quantities of theoretical interest. This style of controlled experimentation---which has appeared in other areas of machine learning research \citep{balakrishnanFace}---might facilitate a richer feedback loop with theory in pursuit of a more complete understanding of generalization.

\subsection{Implications for Learning Theory}
This paper used a controlled study to find a counterexample to a hypothesized causal relationship about generalization in neural networks. This technique could be used in a more positive sense to design improved generalization theories. For instance, one popular framework for generalization theory known as \textit{uniform convergence} derives risk bounds that hold for all classifiers within a specified structural family. Controlled investigation of generalization in different structural families could lead to the discovery of new structural families for uniform convergence theory that are immune to the kind of counterexamples witnessed in this paper.

Controlled experiments may also help in studying other generalization theories such as \textit{PAC-Bayes theory} \citep{mcallester1999some}. PAC-Bayes bounds hold for distributions of classifiers, and controlled studies might enable more efficient investigation of special distributions of classifiers. A concrete example of this is the experiment in Figure \ref{fig:ensembles}, where controlled optimization enables the study of distributions of classifiers conditioned on a prescribed normalized margin.

\subsection{Implications for Uncertainty Quantification}
The techniques developed in this paper may have applications beyond learning theory. In uncertainty quantification, the challenge is to coax a machine learning model into reporting a meaningful notion of confidence in its predictions. One popular technique, known as \textit{deep ensembles} \citep{NIPS2017_9ef2ed4b}, involves training many neural networks with different random seeds in order to obtain a spread of predictions. But according to the Gaussian process model of normalized margin developed in Section \ref{sec:gp-model}, if one is not careful and trains each deep ensemble member to large normalized margin, the trained ensemble members may collapse on to the same function. This model would suggest that to obtain accurate uncertainty information from a deep ensemble, each ensemble member should be trained to small normalized margin. As such, normalized margin control may play a role in uncertainty quantification.

\section*{Acknowledgements}
The authors are grateful to the anonymous reviewers for their helpful comments. This material is based upon work supported by the National Science Foundation Graduate Research Fellowship under Grant No.~DGE‐1745301. This material was also supported by the following grants: NSF \#1918865, ONR N00014-21-1-2483, and a gift from Amazon.

\bibliography{refs}

\begin{thebibliography}{42}
\providecommand{\natexlab}[1]{#1}
\providecommand{\url}[1]{\texttt{#1}}
\expandafter\ifx\csname urlstyle\endcsname\relax
  \providecommand{\doi}[1]{doi: #1}\else
  \providecommand{\doi}{doi: \begingroup \urlstyle{rm}\Url}\fi

\bibitem[Balakrishnan et~al.(2020)Balakrishnan, Xiong, Xia, and
  Perona]{balakrishnanFace}
Balakrishnan, G., Xiong, Y., Xia, W., and Perona, P.
\newblock Towards causal benchmarking of bias in face analysis algorithms.
\newblock In \emph{European Conference on Computer Vision}, 2020.

\bibitem[Bartlett et~al.(2017)Bartlett, Foster, and Telgarsky]{bartlett}
Bartlett, P.~L., Foster, D.~J., and Telgarsky, M.~J.
\newblock Spectrally-normalized margin bounds for neural networks.
\newblock In \emph{Neural Information Processing Systems}, 2017.

\bibitem[Boser et~al.(1992)Boser, Guyon, and Vapnik]{boser_guyon_vapnik}
Boser, B.~E., Guyon, I.~M., and Vapnik, V.~N.
\newblock A training algorithm for optimal margin classifiers.
\newblock In \emph{Workshop on Computational Learning Theory}, 1992.

\bibitem[Cho \& Saul(2009)Cho and Saul]{choandsaul}
Cho, Y. and Saul, L.
\newblock Kernel methods for deep learning.
\newblock In \emph{Neural Information Processing Systems}, 2009.

\bibitem[Cortes \& Vapnik(1995)Cortes and Vapnik]{cortes1995support}
Cortes, C. and Vapnik, V.
\newblock Support-vector networks.
\newblock \emph{Machine Learning}, 1995.

\bibitem[de~G.~Matthews et~al.(2018)de~G.~Matthews, Hron, Rowland, Turner, and
  Ghahramani]{g.2018gaussian}
de~G.~Matthews, A.~G., Hron, J., Rowland, M., Turner, R.~E., and Ghahramani, Z.
\newblock Gaussian process behaviour in wide deep neural networks.
\newblock In \emph{International Conference on Learning Representations}, 2018.

\bibitem[De~Palma et~al.(2019)De~Palma, Kiani, and Lloyd]{de2019random}
De~Palma, G., Kiani, B., and Lloyd, S.
\newblock Random deep neural networks are biased towards simple functions.
\newblock \emph{Neural Information Processing Systems}, 2019.

\bibitem[Dziugaite \& Roy(2017)Dziugaite and Roy]{danroy_nonvacuous}
Dziugaite, G.~K. and Roy, D.~M.
\newblock Computing nonvacuous generalization bounds for deep (stochastic)
  neural networks with many more parameters than training data.
\newblock In \emph{Uncertainty in Artificial Intelligence}, 2017.

\bibitem[Dziugaite et~al.(2020)Dziugaite, Drouin, Neal, Rajkumar, Caballero,
  Wang, Mitliagkas, and Roy]{dziugaite2020search}
Dziugaite, G.~K., Drouin, A., Neal, B., Rajkumar, N., Caballero, E., Wang, L.,
  Mitliagkas, I., and Roy, D.~M.
\newblock In search of robust measures of generalization.
\newblock In \emph{Neural Information Processing Systems}, 2020.

\bibitem[Elsayed et~al.(2018)Elsayed, Krishnan, Mobahi, Regan, and
  Bengio]{elsayed2018large}
Elsayed, G.~F., Krishnan, D., Mobahi, H., Regan, K., and Bengio, S.
\newblock Large margin deep networks for classification.
\newblock In \emph{Neural Information Processing Systems}, 2018.

\bibitem[Geiger et~al.(2020)Geiger, Spigler, Jacot, and Wyart]{Geiger_2020}
Geiger, M., Spigler, S., Jacot, A., and Wyart, M.
\newblock Disentangling feature and lazy training in deep neural networks.
\newblock \emph{Journal of Statistical Mechanics: Theory and Experiment}, 2020.

\bibitem[Herbrich \& Graepel(2001)Herbrich and Graepel]{why-svms-work}
Herbrich, R. and Graepel, T.
\newblock A {PAC-B}ayesian margin bound for linear classifiers: Why {SVM}s
  work.
\newblock In \emph{Neural Information Processing Systems}, 2001.

\bibitem[Hui \& Belkin(2021)Hui and Belkin]{HuiSquareCrossEntropy}
Hui, L. and Belkin, M.
\newblock Evaluation of neural architectures trained with square loss vs.\
  cross-entropy in classification tasks.
\newblock In \emph{International Conference on Learning Representations}, 2021.

\bibitem[Jacot et~al.(2018)Jacot, Gabriel, and Hongler]{NTKjacot}
Jacot, A., Gabriel, F., and Hongler, C.
\newblock Neural tangent kernel: Convergence and generalization in neural
  networks.
\newblock In \emph{Neural Information Processing Systems}, 2018.

\bibitem[Jiang et~al.(2019)Jiang, Krishnan, Mobahi, and
  Bengio]{jiang2018predicting}
Jiang, Y., Krishnan, D., Mobahi, H., and Bengio, S.
\newblock Predicting the generalization gap in deep networks with margin
  distributions.
\newblock In \emph{International Conference on Learning Representations}, 2019.

\bibitem[Jiang et~al.(2020)Jiang, Neyshabur, Mobahi, Krishnan, and
  Bengio]{jiang2019fantastic}
Jiang, Y., Neyshabur, B., Mobahi, H., Krishnan, D., and Bengio, S.
\newblock Fantastic generalization measures and where to find them.
\newblock In \emph{International Conference on Learning Representations}, 2020.

\bibitem[Keskar et~al.(2017)Keskar, Mudigere, Nocedal, Smelyanskiy, and
  Tang]{Keskar2017OnLT}
Keskar, N.~S., Mudigere, D., Nocedal, J., Smelyanskiy, M., and Tang, P. T.~P.
\newblock On large-batch training for deep learning: Generalization gap and
  sharp minima.
\newblock In \emph{International Conference on Learning Representations}, 2017.

\bibitem[Lakshminarayanan et~al.(2017)Lakshminarayanan, Pritzel, and
  Blundell]{NIPS2017_9ef2ed4b}
Lakshminarayanan, B., Pritzel, A., and Blundell, C.
\newblock Simple and scalable predictive uncertainty estimation using deep
  ensembles.
\newblock In \emph{Neural Information Processing Systems}, 2017.

\bibitem[Langford \& Shawe-Taylor(2003)Langford and
  Shawe-Taylor]{langford2003pac}
Langford, J. and Shawe-Taylor, J.
\newblock {PAC-B}ayes \& margins.
\newblock \emph{Neural Information Processing Systems}, 2003.

\bibitem[Lee et~al.(2018)Lee, Sohl-Dickstein, Pennington, Novak, Schoenholz,
  and Bahri]{lee2018deep}
Lee, J., Sohl-Dickstein, J., Pennington, J., Novak, R., Schoenholz, S., and
  Bahri, Y.
\newblock Deep neural networks as {G}aussian processes.
\newblock In \emph{International Conference on Learning Representations}, 2018.

\bibitem[Li et~al.(2018)Li, Xu, Taylor, Studer, and
  Goldstein]{NEURIPS2018_a41b3bb3}
Li, H., Xu, Z., Taylor, G., Studer, C., and Goldstein, T.
\newblock Visualizing the loss landscape of neural nets.
\newblock In \emph{Neural Information Processing Systems}, 2018.

\bibitem[Liu et~al.(2021)Liu, Bernstein, Meister, and Yue]{pmlr-v139-liu21c}
Liu, Y., Bernstein, J., Meister, M., and Yue, Y.
\newblock Learning by turning: Neural architecture aware optimisation.
\newblock In \emph{International Conference on Machine Learning}, 2021.

\bibitem[McAllester(1999)]{mcallester1999some}
McAllester, D.~A.
\newblock Some {PAC-B}ayesian theorems.
\newblock \emph{Machine Learning}, 1999.

\bibitem[Mehta et~al.(2021)Mehta, Cutkosky, and Neyshabur]{extremeMemorization}
Mehta, H., Cutkosky, A., and Neyshabur, B.
\newblock Extreme memorization via scale of initialization.
\newblock In \emph{International Conference on Learning Representations}, 2021.

\bibitem[Mingard et~al.(2021)Mingard, Valle-P{\'e}rez, Skalse, and
  Louis]{mingard2021sgd}
Mingard, C., Valle-P{\'e}rez, G., Skalse, J., and Louis, A.~A.
\newblock Is {SGD} a {B}ayesian sampler? {W}ell, almost.
\newblock \emph{Journal of Machine Learning Research}, 2021.

\bibitem[Nagarajan \& Kolter(2017)Nagarajan and
  Kolter]{nagarajan2019generalization}
Nagarajan, V. and Kolter, J.~Z.
\newblock Generalization in deep networks: The role of distance from
  initialization.
\newblock In \emph{NeurIPS Workshop on Deep Learning: Bridging Theory and
  Practice}, 2017.

\bibitem[Nagarajan \& Kolter(2019)Nagarajan and Kolter]{NEURIPS2019_05e97c20}
Nagarajan, V. and Kolter, J.~Z.
\newblock Uniform convergence may be unable to explain generalization in deep
  learning.
\newblock In \emph{Neural Information Processing Systems}, 2019.

\bibitem[Nakkiran et~al.(2020)Nakkiran, Kaplun, Bansal, Yang, Barak, and
  Sutskever]{Nakkiran2020Deep}
Nakkiran, P., Kaplun, G., Bansal, Y., Yang, T., Barak, B., and Sutskever, I.
\newblock Deep double descent: Where bigger models and more data hurt.
\newblock In \emph{International Conference on Learning Representations}, 2020.

\bibitem[Neal(1994)]{radford}
Neal, R.~M.
\newblock \emph{Bayesian Learning for Neural Networks}.
\newblock {Ph.D.} thesis, Department of Computer Science, University of
  Toronto, 1994.

\bibitem[Neyshabur et~al.(2015)Neyshabur, Tomioka, and
  Srebro]{neyshabur2015norm}
Neyshabur, B., Tomioka, R., and Srebro, N.
\newblock Norm-based capacity control in neural networks.
\newblock In \emph{Conference on Learning Theory}, 2015.

\bibitem[Neyshabur et~al.(2017)Neyshabur, Bhojanapalli, and
  Srebro]{neyshabur2017pac}
Neyshabur, B., Bhojanapalli, S., and Srebro, N.
\newblock A {PAC-B}ayesian approach to spectrally-normalized margin bounds for
  neural networks.
\newblock \emph{International Conference on Learning Representations}, 2017.

\bibitem[P{\'e}rez \& Louis(2020)P{\'e}rez and Louis]{Prez2020GeneralizationBF}
P{\'e}rez, G.~V. and Louis, A.~A.
\newblock Generalization bounds for deep learning.
\newblock \emph{arXiv:2012.04115}, 2020.

\bibitem[Rasmussen \& Williams(2005)Rasmussen and Williams]{gpml}
Rasmussen, C.~E. and Williams, C. K.~I.
\newblock \emph{Gaussian Processes for Machine Learning}.
\newblock MIT Press, 2005.

\bibitem[Rivasplata et~al.(2020)Rivasplata, Kuzborskij, Szepesvári, and
  Shawe-Taylor]{rivasplataKSS20}
Rivasplata, O., Kuzborskij, I., Szepesvári, C., and Shawe-Taylor, J.
\newblock {PAC-B}ayes analysis beyond the usual bounds.
\newblock In \emph{Neural Information Processing Systems}, 2020.

\bibitem[Rosset et~al.(2003)Rosset, Zhu, and Hastie]{rosset2003margin}
Rosset, S., Zhu, J., and Hastie, T.
\newblock Margin maximizing loss functions.
\newblock In \emph{Neural Information Processing Systems}, 2003.

\bibitem[Soudry et~al.(2018)Soudry, Hoffer, Nacson, Gunasekar, and
  Srebro]{soudry2018implicit}
Soudry, D., Hoffer, E., Nacson, M.~S., Gunasekar, S., and Srebro, N.
\newblock The implicit bias of gradient descent on separable data.
\newblock \emph{Journal of Machine Learning Research}, 2018.

\bibitem[Valle-Perez et~al.(2019)Valle-Perez, Camargo, and
  Louis]{valle_perez2018deep}
Valle-Perez, G., Camargo, C.~Q., and Louis, A.~A.
\newblock Deep learning generalizes because the parameter--function map is
  biased towards simple functions.
\newblock In \emph{International Conference on Learning Representations}, 2019.

\bibitem[Vapnik(1999)]{vapnik1999nature}
Vapnik, V.
\newblock \emph{The Nature of Statistical Learning Theory}.
\newblock Springer, 1999.

\bibitem[Woodworth et~al.(2020)Woodworth, Gunasekar, Lee, Moroshko, Savarese,
  Golan, Soudry, and Srebro]{woodworth20a}
Woodworth, B., Gunasekar, S., Lee, J.~D., Moroshko, E., Savarese, P., Golan,
  I., Soudry, D., and Srebro, N.
\newblock Kernel and rich regimes in overparametrized models.
\newblock In \emph{Conference on Learning Theory}, 2020.

\bibitem[Wu et~al.(2017)Wu, Zhu, and Weinan]{Wu2017TowardsUG}
Wu, L., Zhu, Z., and Weinan, E.
\newblock Towards understanding generalization of deep learning: Perspective of
  loss landscapes.
\newblock In \emph{ICML Workshop on Principled Approaches to Deep Learning},
  2017.

\bibitem[Zhang et~al.(2021{\natexlab{a}})Zhang, Bengio, Hardt, Recht, and
  Vinyals]{zhang2021understanding}
Zhang, C., Bengio, S., Hardt, M., Recht, B., and Vinyals, O.
\newblock Understanding deep learning (still) requires rethinking
  generalization.
\newblock \emph{Communications of the ACM}, 2021{\natexlab{a}}.

\bibitem[Zhang et~al.(2021{\natexlab{b}})Zhang, Reid, Pérez, and
  Louis]{zhang2021flatness}
Zhang, S., Reid, I., Pérez, G.~V., and Louis, A.
\newblock Why flatness does and does not correlate with generalization for deep
  neural networks.
\newblock \emph{arXiv:2103.06219}, 2021{\natexlab{b}}.

\end{thebibliography}
\bibliographystyle{icml2022/icml2022}

\clearpage
\appendix
\section{Experimental Details}

\subsection{Reversing Spectrally Normalized Margin Bounds}
\label{sec:spectral_ap}
Depth 5, width 5000 fully connected neural networks were trained for 10-class classification on subsets of 1000 training points from MNIST and evaluated on the entire MNIST test set.  Rectified Linear unit (ReLU) activations were used throughout all experiments.  The data was normalized according to Recipe \ref{alg:control} without the Frobenius control.  The networks generating the margin distributions in Figure \ref{fig:bartlett_reproduce} were trained with a label-scaled squared loss function (true data label scale: 1, random data label scale: 10), full batch gradient descent with a learning rate of 0.01 and and an exponential learning rate decay of 0.999.  They were trained to within 95\% training accuracy.  The networks in Figure \ref{fig:bartlett_reproduce} had test performance of 88\% for the correctly labeled network and 8.3\% for the randomly labeled network.

The networks generating the margin distributions in Figure \ref{fig:bartlett_placeholder} had identical architectures as above, but were trained with Frobenius control using the Nero optimizer (learning rate: 0.01, Nero $\beta$: 0.999) to perfect classification accuracy.  Targeted label scales were set to 1 (true data) and 100 (random labels).  Spectral-normalization was calculated with respect to the weights at initialization and included $\|X\|$ and $n$ correction.  The networks in this figure had test performance of 81\% for the correctly labeled network and 10\% for the randomly labeled network.

\subsection{Twin Network Study: Attack Set}
\label{sec:attack_set_ap}

\begin{figure}
\centering
\includegraphics[width=\columnwidth]{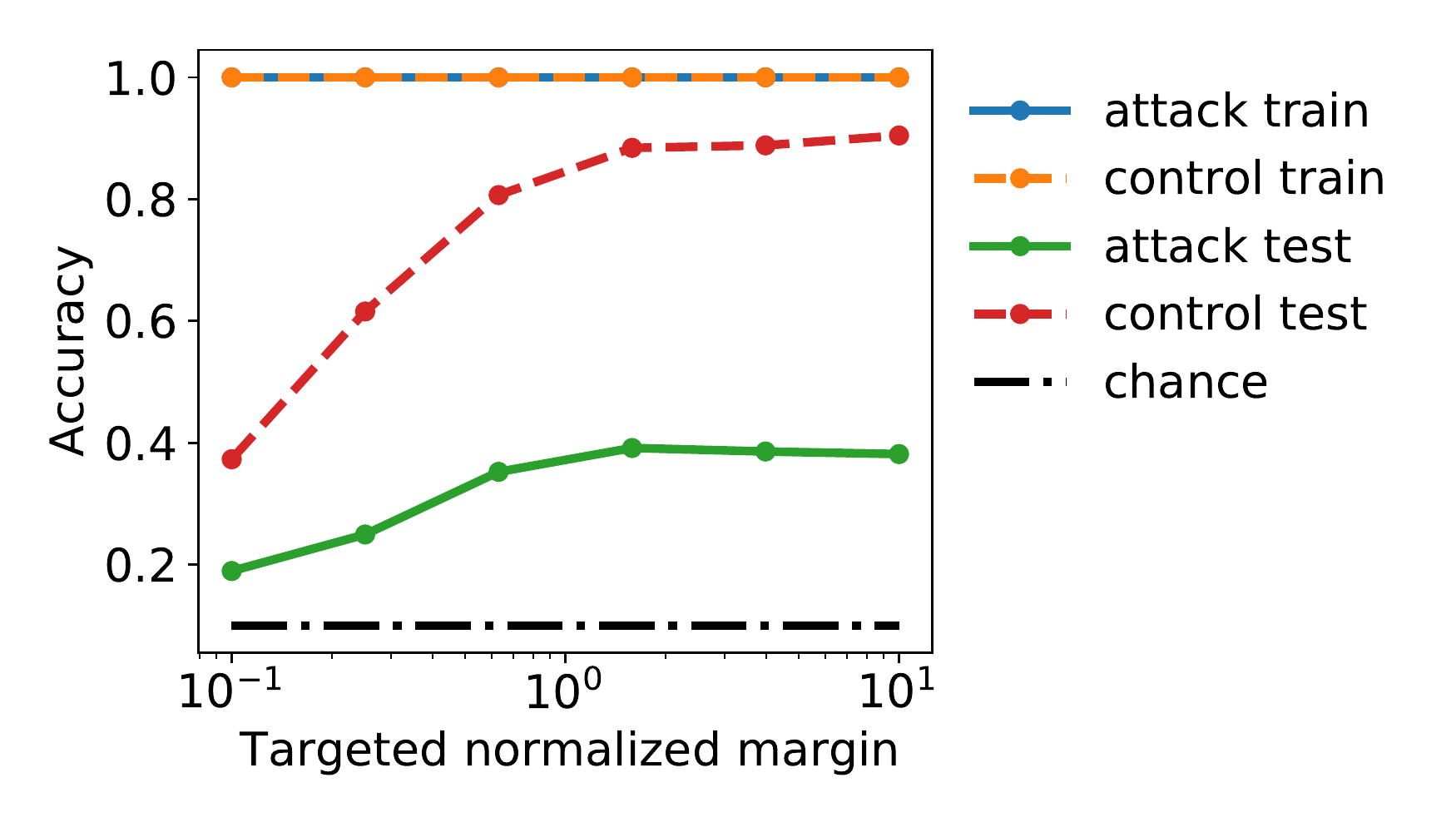}
\caption{Test and training accuracy of twin networks with or without addition of an attack set as a function of targeted Frobenius normalized margin.}
\label{fig:attack_summary}
\end{figure}

Two layer fully connected neural networks (width: 2048) were trained using Frobenius control with Nero (learning rate: 0.01, $\beta$: 0.99997, 100,000 epochs) to fit 500 training points from MNIST for 10-class classification.  One network's training set was further augmented by adding the attack set of the 1000 more train points labeled randomly.  They were both evaluated on the correctly labeled 10,000 test points and achieved perfect classification accuracy.  Only the margins for the correctly labeled 500 training points are presented in Figure \ref{fig:augment_hist_placeholder}.  Figure \ref{fig:attack_summary} shows accuracy of twin networks that have (attack) or have not (control) been trained on an attack set as a function of targeted normalized margin.

\subsection{Twin Network Study: Optimization Dependence}
\label{sec:twin_network_exp_details}

For MNIST experiments, the architecture was a depth 7 MLP with ReLU activation. For CIFAR-10 experiments, the architecture consisted of 3 convolutional layers, followed by a flatten, followed by 3 linear layers. For MNIST experiments, the intermediate layer widths used were: 500, 1000, 2000, 3000, 4000, 5000, 6000, 7000. For CIFAR-10 experiments, the width of the linear layers were fixed to be 500, and the channel width of the convolutional layers varied as per the following list: 3, 5, 10, 20, 40, 80, 160, 320. When performing random sampling of parameters, weights were drawn from a $\mathcal{N}(0, 1)$ distribution.

To ensure the chosen architectures could capture their binary classification task, all architectures were trained on their respective binary classification task for 50 epochs. For MNIST 0 vs. 1 classification, the training set size was 12665 and test set size was 2115. For MNIST 4 vs. 7 classification, the training set size was 12107 and test size was 2010. For MNIST 3 vs. 8 classification, the training set size was 11982 and test set size was 1984. For CIFAR-10 dog vs. ship, the training set size was 10000 and test set size was 2000. On MNIST binary classification tasks (0 vs. 1, 3 vs. 8, and 4 vs. 7), the worst training accuracy across all architectures at the end of training was 100\%; for CIFAR-10, the worst training accuracy across all architectures at the end of training was 73.81\%.

As noted in section \ref{sec:attack}, networks were trained to match the margin of sampled networks.  Training used a loss threshold of 0.000001, which indicates that $||f_{\text{sampled}}(x) - f_{\text{trained}}(x)||_2 <$ 0.000001 for the given set of training examples $x$. To justify this choice, training loss was inspected relative to the scale of the margin of $f_{\text{sampled}}$ for each pair $(f_{\text{trained}}, f_{\text{sampled}})$.  Table \ref{table:svt_experimental_details} reports the worst relative error across all architectures and seeds for a corresponding binary classification task.  The worst relative error is very small, indicating that there is a negligible difference in margin between $f_{\text{sampled}}$ and $f_{\text{trained}}$.

\subsection{Normalized Margin in Standard Training}
\label{sec:normalized_margin_ap}
\begin{figure}
\centering
\includegraphics[width=\columnwidth]{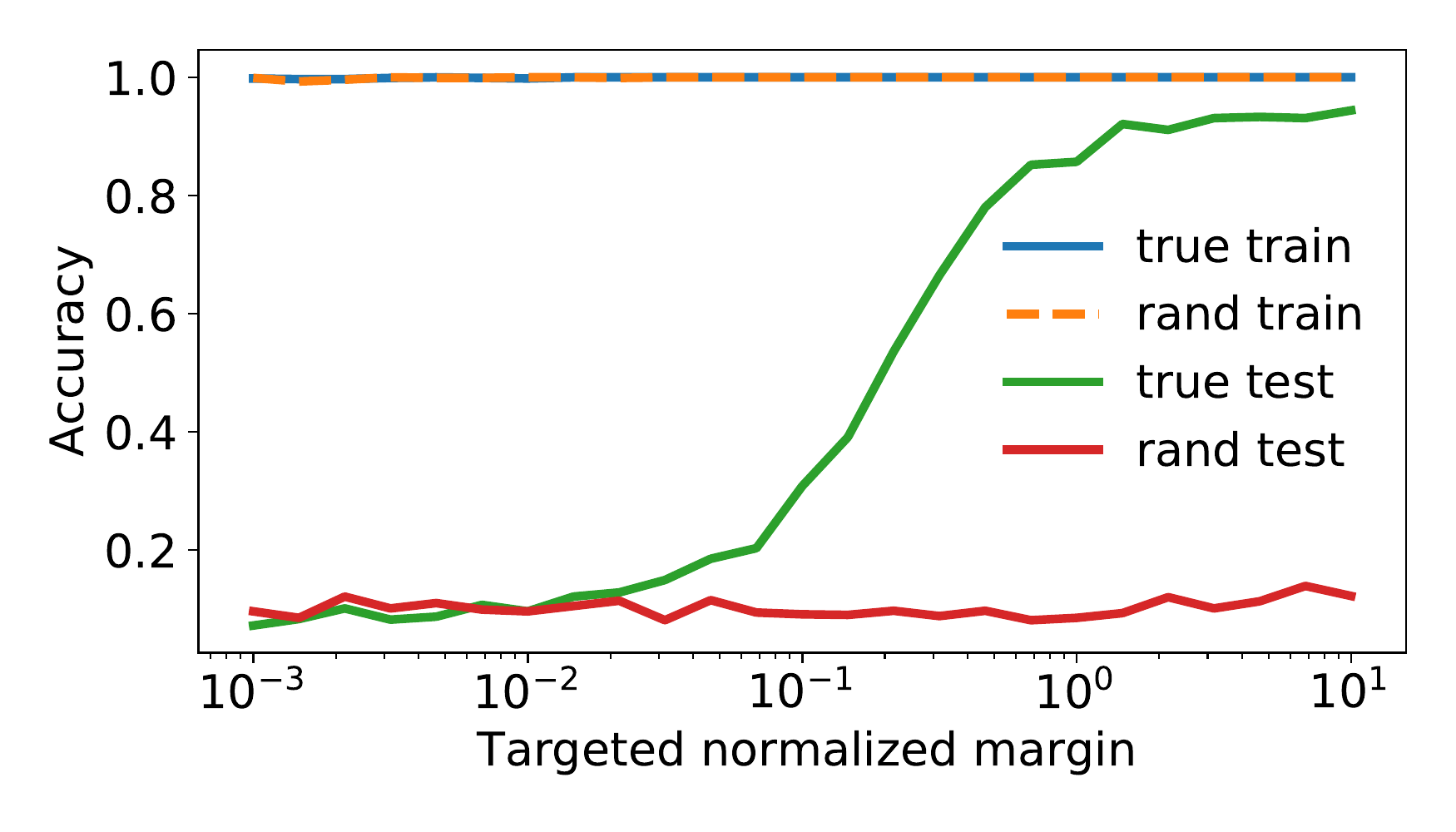}
\caption{Train and test performance of networks trained to target specified Frobenius-normalized margins.}
\label{fig:nero_em_train_vs_rand}
\end{figure}

2-Layer neural networks were trained to fit 1000 point subsets of MNIST and evaluated on the whole test set.  They were trained using either full batch gradient descent or full batch Nero ($\beta$: 0.999) while varying the initialization scale or targeted margin scale.  Networks were trained between 50,000 to 250,000 epochs (learning rates between 0.9998 and 0.999998) to achieve training accuracy marked in Figure \ref{fig:em_placeholder}.  Figure \ref{fig:nero_em_train_vs_rand} shows accuracy as a function of Frobenius normalized targeted margin for networks trained on true or random data.

\subsection{Ensemble Behavior of Normalized Margin}
\label{sec:ensembles_set_ap}

Depth 5, width 2048 MLPs were trained on 1000 samples of MNIST digits, to perform even/odd classification. Networks were trained using full-batch Nero with initial learning rate 0.01, beta set to 0.999 and learning rate decay factor 0.99 per iteration. Networks were trained for 500 iterations. A variety of margins were targeted ranging from $10^{-3}$ up to $10^{1}$. The experiment was repeated for Gaussian process draws using the predictive given in Equation \ref{eq:gp-ensemble} with $C_1$ and $C_2$ defined earlier in that section.

\newpage

\begin{figure}
\centering
\includegraphics[width=\columnwidth]{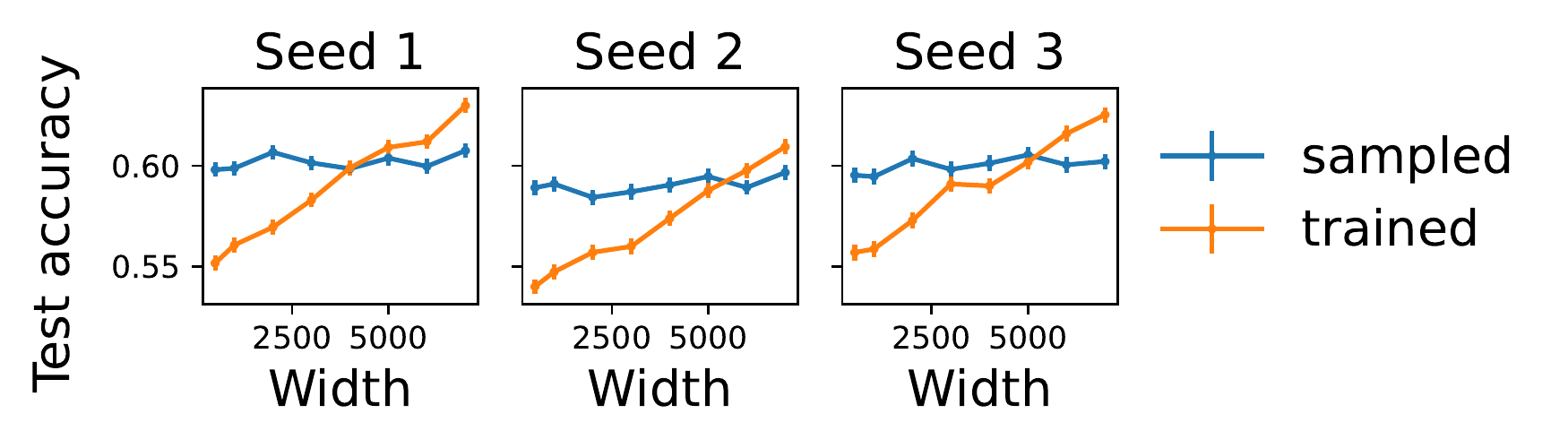}
\includegraphics[width=\columnwidth]{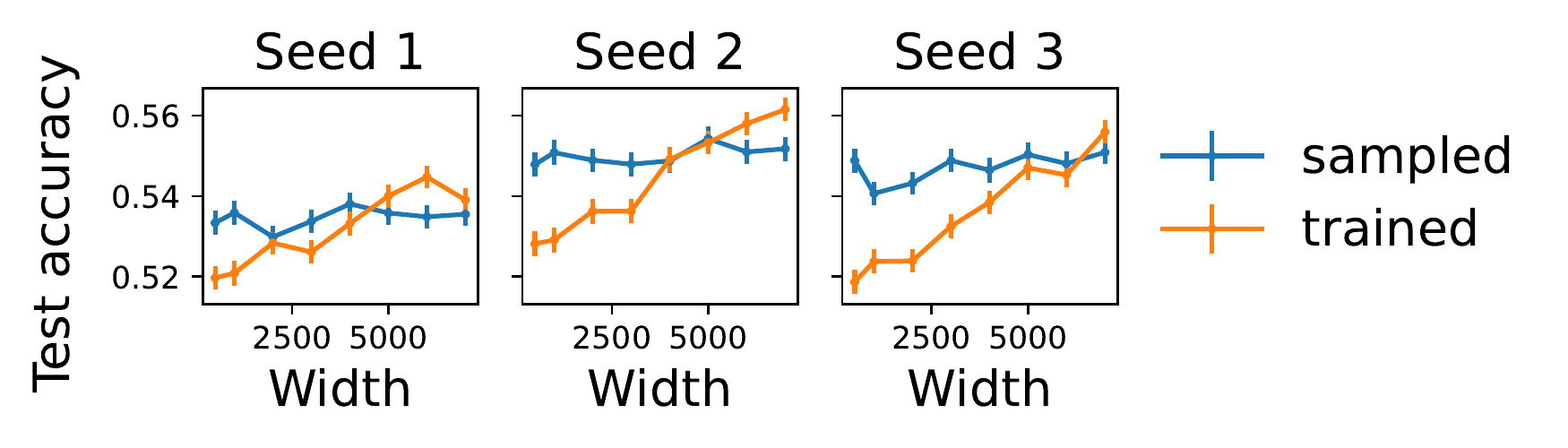}
\includegraphics[width=\columnwidth]{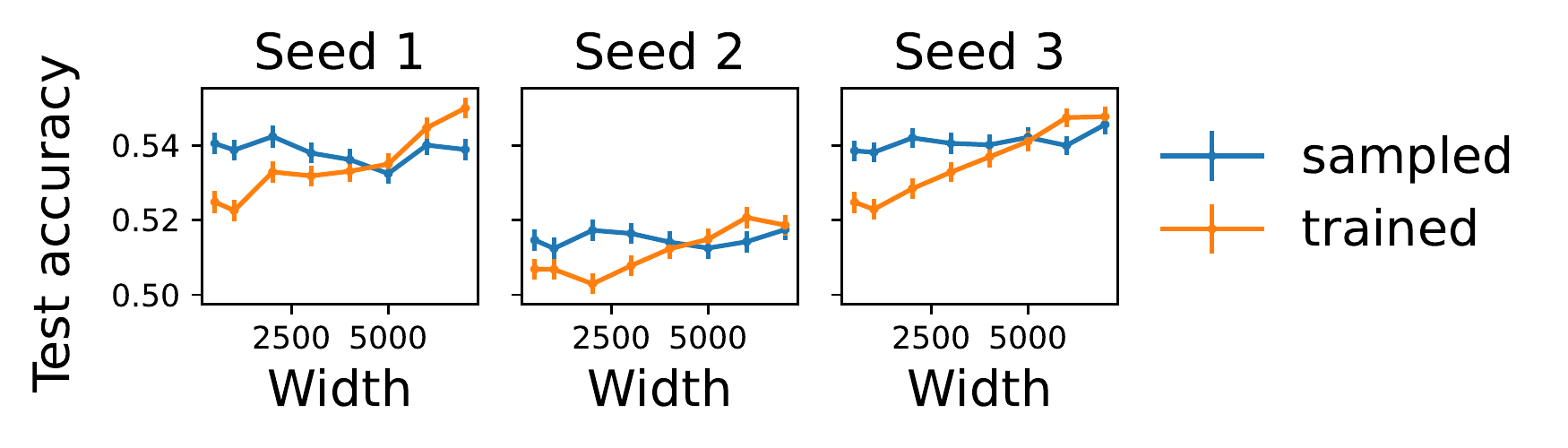}
\caption{Test performance of $f_{\text{sampled}}$ vs $f_{\text{trained}}$ for 0 vs. 1 (top), 4 vs. 7. (middle), and 3 vs. 8 (bottom) MNIST binary classification task. For each width, the mean test accuracy and standard error bars are presented for $1000$ pairs of ($f_{\text{sampled}}$, $f_{\text{trained}}$).}
\label{fig:mnist_svt_full_seeds}
\end{figure}

\begin{figure}
\centering
\includegraphics[width=\columnwidth]{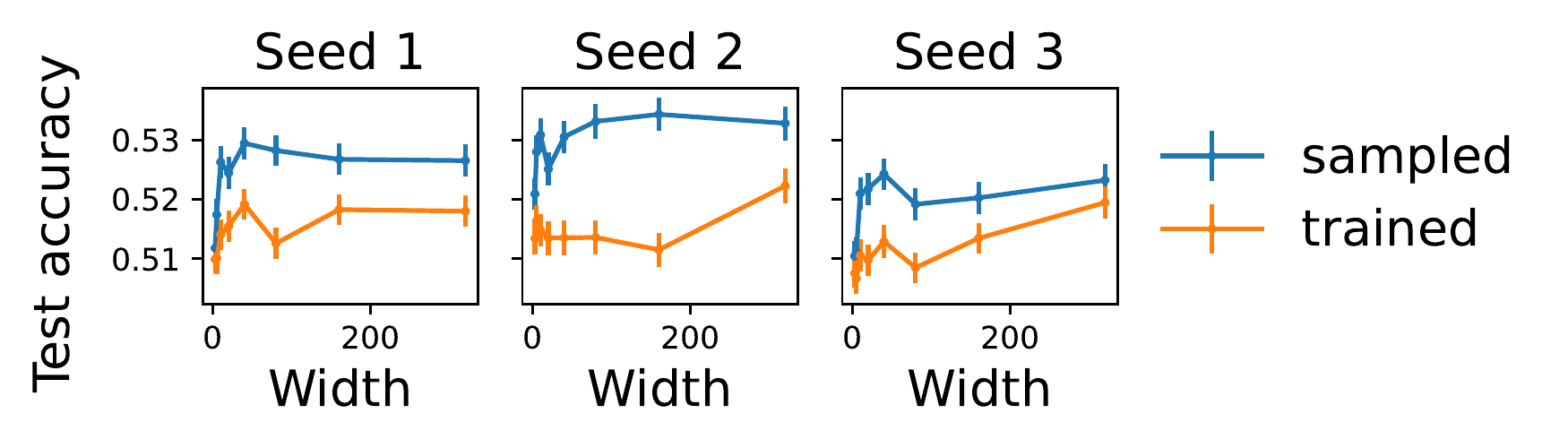}
\caption{Test performance of $f_{\text{sampled}}$ vs $f_{\text{trained}}$ for dog vs. ship CIFAR-10 binary classification task. For each width, the mean test accuracy and standard error bars are presented for $1000$ pairs of ($f_{\text{sampled}}$, $f_{\text{trained}}$).}
\label{fig:cifar_dogvship_five_seeds}
\end{figure}

\begin{table}
\centering
\begin{tabular}{c c}
\toprule
\textbf{Learning task} & \textbf{Worst relative error} \\
\midrule
MNIST: 0 vs. 1 &  0.0000355 \\
MNIST: 3 vs. 8 & 0.0000568 \\
MNIST: 4 vs. 7 & 0.0000248 \\
CIFAR-10: dog vs. ship & 0.0002589 \\
\bottomrule
\end{tabular}
\caption{Worst relative error between randomly sampled and gradient-descent trained networks across all 1000 samples and seeds, for a given learning task.}
\label{table:svt_experimental_details}
\end{table}

\end{document}